\documentclass[letterpaper,10pt]{article}

\usepackage{probml}

\ShortHeadings{An Isotropic Approach to Efficient Uncertainty Quantification with Gradient Norms}

\usepackage{multirow}
\usepackage{wrapfig}
\usepackage{microtype}
\usepackage{nicefrac}
\usepackage{subcaption}

\usepackage{bm,amssymb,amsmath}

\makeatletter
\def\th@plain{%
  \thm@notefont{}
  \itshape 
}
\def\th@definition{%
  \thm@notefont{}
  \normalfont 
}
\makeatother

\def\1{\bm{1}}

\DeclareMathAlphabet{\mathsfit}{\encodingdefault}{\sfdefault}{m}{sl}
\SetMathAlphabet{\mathsfit}{bold}{\encodingdefault}{\sfdefault}{bx}{n}

\begin{document}

\title{An Isotropic Approach to Efficient Uncertainty Quantification with Gradient Norms}

\author[1,3,$\dagger$]{Nils Gr{\"u}nefeld}
\author[2,3]{Jes Frellsen}
\author[1,3]{Christian Hardmeier}

\affil[1]{IT University of Copenhagen}
\affil[2]{Technical University of Denmark}
\affil[3]{Pioneer Centre for Artificial Intelligence}
\affil[$\dagger$]{Correspondence to \url{nilgr@itu.dk}}

\maketitle

\begin{abstract}
Existing methods for quantifying predictive uncertainty in neural networks are either computationally intractable for large language models or require access to training data that is typically unavailable.
We derive a lightweight alternative through two approximations: a first-order Taylor expansion that expresses uncertainty in terms of the gradient of the prediction and the parameter covariance, and an isotropy assumption on the parameter covariance.
Together, these yield epistemic uncertainty as the squared gradient norm and aleatoric uncertainty as the Bernoulli variance of the point prediction, from a single forward-backward pass through an unmodified pretrained model.
We justify the isotropy assumption by showing that covariance estimates built from non-training data introduce structured distortions that isotropic covariance avoids, and that theoretical results on the spectral properties of large networks support the approximation at scale.
Validation against reference Markov Chain Monte Carlo estimates on synthetic problems shows strong correspondence that improves with model size.
We then use the estimates to investigate when each uncertainty type carries useful signal for predicting answer correctness in question answering with large language models, revealing a benchmark-dependent divergence: the combined estimate achieves the highest mean AUROC on TruthfulQA, where questions involve genuine conflict between plausible answers, but falls to near chance on TriviaQA's factual recall, suggesting that parameter-level uncertainty captures a fundamentally different signal than self-assessment methods.
\end{abstract}

\section{Introduction}
\label{sec:introduction}

As large language models (LLMs) are increasingly deployed in consequential applications---from medical diagnosis assistance to legal document analysis and financial advisory services \citep{chen-2024-surveylarge}---ensuring their trustworthiness becomes paramount.
A fundamental requirement is the ability to assess when a model's predictions may be unreliable, yet contemporary LLMs provide no built-in mechanism for distinguishing what they know from what they do not, invariably delivering outputs with the same authoritative tone regardless of whether their training provides adequate foundation for the claims they make \citep{ji-2023-surveyhallucination, zhou-2024-relyingunreliable}.

Unreliable predictions can arise for two fundamentally different reasons.
Some predictions are uncertain because the question itself admits multiple valid answers, an inherent ambiguity that no amount of additional training data would resolve.
Others are uncertain because the model has seen too little relevant training data to have settled on a reliable answer, a gap in knowledge that more data could, in principle, fill.
These two sources are known as \emph{aleatoric} and \emph{epistemic} uncertainty, respectively, and distinguishing between them is essential: the former signals an irreducibly hard problem, while the latter flags a prediction that should not be trusted.

The Bayesian framework provides a natural language for quantifying both sources of uncertainty \citep{neal-2012-bayesianlearning, kendall-2017-whatuncertainties}, but applying it is intractable for modern neural networks \citep{blundell-2015-weightuncertainty}, and existing approximations---deep ensembles \citep{lakshminarayanan-2017-simplescalable}, Monte Carlo dropout \citep{gal-2016-dropoutbayesian}, Laplace approximations \citep{mackay-1992-practicalbayesian, daxberger-2021-laplaceredux}---each impose severe practical limitations: training multiple models, requiring architectural support, or computing Hessian matrices from training data that is typically unavailable for contemporary LLMs.

We address this gap through two approximations.
First, a first-order Taylor expansion, also known as the delta method \citep{doob-1935-limitingdistributions}, expresses uncertainty as a product of two factors: the gradient of the prediction with respect to the parameters, and the covariance of those parameters.
Second, we assume isotropic parameter covariance, leaving only the gradient factor as our epistemic estimate.
The same expansion yields the aleatoric estimate directly from the model's output, giving a complete uncertainty decomposition from a single forward-backward pass through an unmodified pretrained model, with no ensembles, no sampling, and no Hessian estimation.

The isotropy assumption is the key simplification: we argue that proxy covariance estimates built from non-training data can introduce structured distortions worse than isotropy's uniform error, supported by theoretical results on Hessian spectra at scale and empirical precedent across adjacent fields.
We validate against reference MCMC estimates on synthetic problems (Spearman $\rho$ of $0.44$--$0.99$, improving with model scale as the theory predicts) and then investigate when aleatoric and epistemic uncertainty carry useful signal for predicting answer correctness in LLM question answering, revealing a benchmark-dependent divergence: the combined estimate achieves the highest area under the receiver operating characteristic curve (AUROC) ($0.63$) on TruthfulQA, while epistemic uncertainty falls to near chance on TriviaQA, suggesting that parameter-level uncertainty is most informative when the task involves genuine conflict between plausible answers rather than factual memorization.

Our contributions are as follows:
\begin{enumerate}
    \item We derive epistemic and aleatoric uncertainty estimators from a first-order Taylor expansion under an isotropy assumption, providing the first systematic justification for the isotropy assumption through proxy bias analysis, spectral theory, and empirical precedent.
    \item We validate both estimators against reference Markov Chain Monte Carlo (MCMC) estimates on synthetic problems, demonstrating strong correspondence in classification and an improving trend with model scale.
    \item We investigate the utility of aleatoric and epistemic uncertainty for predicting answer correctness in LLM question answering, revealing a benchmark-dependent divergence that illuminates when parameter-level uncertainty provides useful signal.
\end{enumerate}

\section{Background}
\label{sec:bg}

\subsection{Uncertainty in Machine Learning}
\label{sec:bg:uncertainty}

The Bayesian framework reasons about uncertainty by maintaining a posterior distribution $p(\theta \mid \mathcal{D})$ over model parameters $\theta$ given data $\mathcal{D}$ rather than a point estimate; predictions for a new input $x$ are made by marginalizing over this posterior, $p(y \mid x) = \mathbb{E}_\theta[p(y \mid x, \theta)]$, where $y$ denotes the output.
The uncertainty in this predictive distribution decomposes into two components \citep{hullermeier-2021-aleatoricepistemic}: \emph{aleatoric uncertainty}, capturing irreducible randomness in the data-generating process, and \emph{epistemic uncertainty}, reflecting incomplete knowledge about the parameters due to finite training data.
Regularized loss minimization is mathematically equivalent to maximum a posteriori (MAP) estimation \citep{bishop-2006-patternrecognition, goodfellow-2016-deeplearning}, so the loss surface of any regularized neural network is a posterior parameter distribution; standard training collapses this distribution to its mode, discarding the information necessary for uncertainty quantification.

\subsection{Uncertainty Decomposition}
\label{sec:bg:decomposition}

The most widely used decomposition operates on the entropy of the predictive distribution \citep{smith-2018-understandingmeasures}:
\begin{equation}
\underbrace{\mathbb{H}[y \mid x]}_{\text{total}} = \underbrace{\mathbb{E}_\theta\!\left[\mathbb{H}[y \mid x, \theta]\right]}_{\text{aleatoric}} + \underbrace{\mathbb{I}[y; \theta \mid x]}_{\text{epistemic}},
\label{eq:entropy-decomp}
\end{equation}
where the aleatoric component is the expected conditional entropy and the epistemic component is the mutual information between the prediction and the parameters \citep{gal-2016-dropoutbayesian, kuhn-2022-semanticuncertainty}.
However, \citet{wimmer-2023-quantifyingaleatoric} show that mutual information violates several desirable axiomatic properties---it is not maximal under complete ignorance, not monotone under mean-preserving spreads, and not invariant under location shifts---and existing estimators degrade to near-random performance under non-trivial aleatoric uncertainty \citep{tomov-2025-entropynot}.

An alternative decomposition considers variance rather than entropy and operates label-wise \citep{sale-2023-secondorderuncertainty, sale-2024-labelwisealeatoric}.
For a given class $c$, the model's prediction $p(y_c \mid x, \theta)$ can be viewed as the success probability of a Bernoulli variable.
Applying the law of total variance to the posterior gives:
\begin{equation}
\underbrace{\operatorname{Var}\!\left[y_c \mid x\right]}_{\text{total}} = \underbrace{\mathbb{E}_\theta\!\left[p(y_c \mid x, \theta)(1 - p(y_c \mid x, \theta))\right]}_{\text{aleatoric}} + \underbrace{\operatorname{Var}_\theta\!\left[p(y_c \mid x, \theta)\right]}_{\text{epistemic}}.
\label{eq:variance-decomp}
\end{equation}
This label-wise decomposition provides a more fine-grained view than the entropy-based one and does not suffer from the axiomatic violations above \citep{sale-2024-labelwisealeatoric}; we adopt it throughout this work.

Both decompositions assume that the prediction at the MAP estimate equals the posterior predictive expectation, $p(y \mid x, \theta^*) = \mathbb{E}_\theta[p(y \mid x, \theta)]$, a necessary condition for both the law of total variance and the entropy identity.
This assumption is incorrect in practice, but these decompositions remain a standard analytical tool \citep{smith-2018-understandingmeasures, depeweg-2018-decompositionuncertainty, jayasekera-2025-variationaluncertaintya}.
\section{Gradient-Based Epistemic Uncertainty Quantification}
\label{sec:method}

We approximate the predictive distribution via a first-order Taylor expansion around the parameter point estimate $\theta^*$: $p(y_c \mid x, \theta) \approx p(y_c \mid x, \theta^*) + g^{\top}(\theta - \theta^*)$, where $g = \nabla_\theta p(y_c \mid x, \theta)\big|_{\theta^*}$.
Substituting into the variance-based definition of epistemic uncertainty, $\operatorname{Var}_\theta\!\left[p(y_c \mid x, \theta)\right]$:
\begin{align}
\operatorname{Var}_\theta\!\left[p(y_c \mid x, \theta)\right]
&\approx \operatorname{Var}_\theta\!\left[
p(y_c \mid x, \theta^*)
+ g^{\top}(\theta - \theta^*)
\right] \nonumber \\
&= \operatorname{Var}_\theta\!\left[g^{\top}(\theta - \theta^*)\right]
= \operatorname{Var}_\theta\!\left[g^{\top}\theta\right] \nonumber \\
&= g^{\top}\operatorname{Cov}[\theta]\, g
\label{eq:taylor}
\end{align}
This is known as the delta method.
However, at this point we are still left with a notoriously difficult object: the parameter covariance matrix.
Most work on the delta method accordingly focuses on different estimations of this matrix \citep{nilsen-2022-epistemicuncertainty, schmitt-2025-generaluncertainty}.
In contrast, we take our approximation a step further by assuming isotropic parameter covariance.
Since any isotropic covariance $\sigma^2 I$ differs from the identity only by a constant factor that scales all estimates equally, we can set $\operatorname{Cov}[\theta] = I$ without loss of generality:
\begin{align}
\operatorname{Var}_\theta\!\left[p(y_c \mid x, \theta)\right]
\approx g^{\top}\operatorname{Cov}[\theta]\, g
\approx g^{\top} g
\end{align}
We are left with $\|g\|^2$, the squared gradient norm, as our estimate of epistemic uncertainty.
This is a strong assumption that requires justification.

We motivate the isotropy assumption from three complementary angles: (i) the available alternatives to estimate the true covariance may introduce structured distortions worse than assuming isotropy (\cref{sec:method:idcov:proxy}), (ii) theoretical results on the Hessian spectrum suggest that the identity is a reasonable proxy for the true covariance as model size grows, and (iii) empirical precedent confirms that it performs well across tasks adjacent to uncertainty quantification (\cref{sec:method:idcov:theory}).

\subsection{Proxy Covariance Estimates Introduce Structured Bias}
\label{sec:method:idcov:proxy}

\begin{figure}[h]
    \centering
    \begin{subfigure}[t]{0.24\textwidth}
        \centering
        \includegraphics[width=\textwidth]{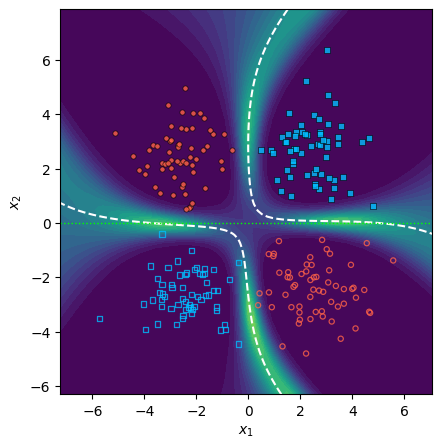}
        \caption{Identity ($C = I$)}
    \end{subfigure}
    \hfill
    \begin{subfigure}[t]{0.24\textwidth}
        \centering
        \includegraphics[width=\textwidth]{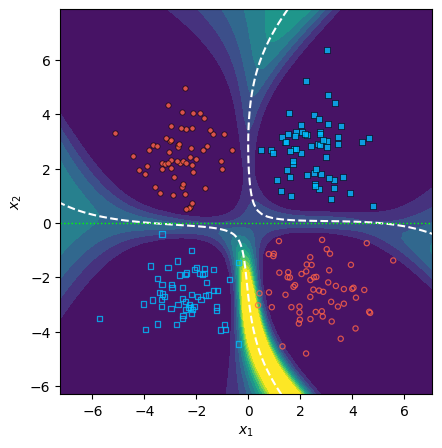}
        \caption{Proxy $H_A^{-1}$ (top half)}
    \end{subfigure}
    \hfill
    \begin{subfigure}[t]{0.24\textwidth}
        \centering
        \includegraphics[width=\textwidth]{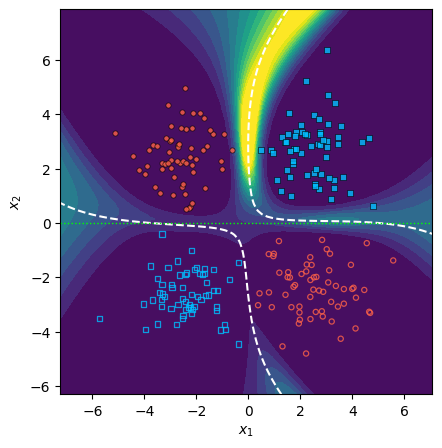}
        \caption{Proxy $H_B^{-1}$ (bottom half)}
    \end{subfigure}
    \hfill
    \begin{subfigure}[t]{0.24\textwidth}
        \centering
        \includegraphics[width=\textwidth]{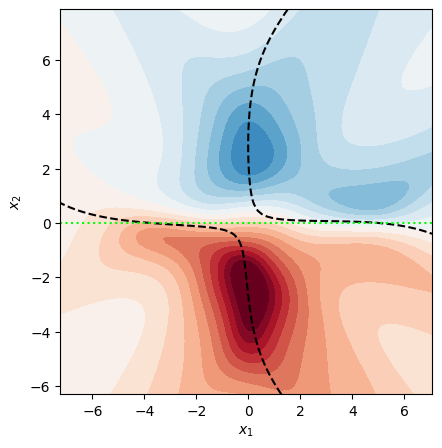}
        \caption{$\log(U_A / U_B)$}
    \end{subfigure}
    \caption{Normalized epistemic uncertainty on the XOR problem under three covariance assumptions, and the log ratio $\log(U_A / U_B)$.
    The identity produces spatially symmetric estimates that peak along the two decision boundaries.
    Each proxy inflates uncertainty in the half of input space \emph{absent} from its data and suppresses it where its data is concentrated, despite the problem's underlying symmetry; the log ratio makes this asymmetry explicit.
    See \cref{app:hessian-bias} for experimental details and additional problems.}
    \label{fig:hessian-bias}
\end{figure}

For LLMs, the true training data is typically unavailable even for ostensibly open models \citep{touvron-2023-llama2, grattafiori-2024-llama3, abdin-2024-phi4technical, team-2025-gemma3}, so any covariance estimate must be built from proxy data: corpora that plausibly overlap with the true training distribution but inevitably do not match it exactly.

Consider a diagonal Laplace approximation \citep{daxberger-2021-laplaceredux}, the closest widely used alternative to isotropic covariance \citep{denker-1990-transformingneuralnet, gui-2021-laplaceapproximation, ortega-2024-variationallinearized}.
Let $\Sigma_{\mathrm{diag}}^*$ denote the true diagonal of $\Sigma$ and $\hat{\Sigma}_{\mathrm{diag}}$ the diagonal estimated from proxy data.
The total error decomposes as
\begin{equation}
g^\top \Sigma\, g - g^\top \hat{\Sigma}_{\mathrm{diag}}\, g
= \underbrace{g^\top (\Sigma - \Sigma_{\mathrm{diag}}^*)\, g}_{\text{(i) structural error}}
\;+\; \underbrace{g^\top (\Sigma_{\mathrm{diag}}^* - \hat{\Sigma}_{\mathrm{diag}})\, g}_{\text{(ii) estimation error}},
\label{eq:diaglaplace-decomp}
\end{equation}
and identically for our isotropic approximation:
\begin{equation}
g^\top \Sigma\, g - \|g\|^2
= \underbrace{g^\top (\Sigma - \Sigma_{\mathrm{diag}}^*)\, g}_{\text{(i) structural error}}
\;+\; \underbrace{g^\top (\Sigma_{\mathrm{diag}}^* - I)\, g}_{\text{\makebox[0pt]{\hss(iii) anisotropy error\hss}}}.
\label{eq:identity-decomp}
\end{equation}
Since the structural error is identical, the difference reduces to comparing terms~(ii) and~(iii):
\begin{equation}
\text{(ii)} = \sum_{i=1}^{P} g_i^2 ( \Sigma_{ii} - \hat{\Sigma}_{ii} ),
\qquad
\text{(iii)} = \sum_{i=1}^{P} g_i^2 ( \Sigma_{ii} - 1 ).
\label{eq:diagonal-comparison}
\end{equation}
Both errors are weighted by $g_i^2$, but the proxy biases $\Sigma_{ii} - \hat{\Sigma}_{ii}$ reflect the coverage of the proxy corpus: parameters that the proxy data activates receive large curvature and small variance, while parameters it does not activate retain poorly constrained variance.
Since $g_i^2$ upweights the parameters the model relies on for a given prediction, the proxy's errors concentrate precisely where accuracy matters most.
The identity's biases $\Sigma_{ii} - 1$ encode no data-dependent structure and therefore cannot introduce spatially structured distortions of this kind.

We demonstrate this empirically on three 2D classification problems with spatially symmetric decision boundaries (details in \cref{app:hessian-bias}).
Splitting the training data by location and computing empirical proxies of the Fisher information matrix (FIM) from each half yields two Hessian estimates $H_A$ and $H_B$, with corresponding uncertainty estimates $U_A(x) = g^\top H_A^{-1} g$ and $U_B(x) = g^\top H_B^{-1} g$.
As \cref{fig:hessian-bias} shows for the XOR (exclusive or) problem, each proxy inflates uncertainty in the half of input space absent from its data and suppresses it where its data is concentrated (Cohen's $d = -2.5$), while the identity produces spatially symmetric estimates that peak at the decision boundary, as the problem's symmetry demands.

\begin{wraptable}{r}{0.50\textwidth}
    \centering
    \small
    \vspace{-1em}
    \begin{tabular}{lccccc}
        \toprule
        & \multicolumn{4}{c}{Test domain} & \\
        \cmidrule(lr){2-5}
        Covariance & \texttt{sci} & \texttt{comp} & \texttt{rec} & \texttt{talk} & CV \\
        \midrule
        representative   & 0.99 & 1.05 & 1.16 & 0.80 & 0.13 \\
        identity (ours)  & 0.83 & 1.25 & 0.75 & 1.17 & 0.21 \\
        \texttt{sci}     & 0.66 & 1.15 & 0.72 & 1.46 & 0.33 \\
        \texttt{rec}     & 0.84 & 1.54 & 0.47 & 1.15 & 0.40 \\
        \texttt{talk}    & 1.01 & 1.88 & 0.78 & 0.32 & 0.57 \\
        \texttt{comp}    & 0.65 & 0.28 & 1.18 & 1.88 & 0.60 \\
        \bottomrule
    \end{tabular}
    \caption{Mean uncertainty per test-sample domain divided by the full Hessian's mean uncertainty on the same domain, normalized by each row's mean. CV: coefficient of variation, computed before row-normalization.}
    \label{tab:hessian-bias-bert-cv-main}
\end{wraptable}

To confirm this beyond synthetic settings, we fine-tune a single 20-class DistilBERT \citep{sanh-2020-distilbertdistilled} on all of 20~Newsgroups and compute proxy Hessians from four topical subsets (\texttt{sci}, \texttt{comp}, \texttt{rec}, \texttt{talk}), as well as a representative sample from all four and a pooled full Hessian (details in \cref{app:hessian-bias-bert}).
Comparing each covariance against the full Hessian on a shared test set, the identity preserves the uncertainty ranking better than any single-domain proxy (Spearman $\rho = 0.979$ vs. $0.920$--$0.962$).

Dividing each method's per-domain ratio to the full Hessian by its own mean (so $1.00$ would indicate a pure global scale factor), every single-domain proxy has its lowest value on the domain its proxy data covers, systematically suppressing uncertainty there.
The identity's deviations are non-zero but smaller and not aligned with any particular domain: its coefficient of variation across domains ($0.21$) is closer to that of the representative sample ($0.13$) than to any single-domain proxy ($0.33$--$0.60$).
Sweeping over all $14$ ways to combine the four domain groups, adding domains tends to reduce CV but not reliably: only $3$ of the $10$ multi-domain proxies fall below the identity's $0.21$, and one triple has CV $=0.54$ (\cref{tab:hessian-bias-bert-combos}).
A practitioner without access to the training data has no way to predict which combinations will be safe.

In short, any covariance estimate for an LLM must be built from proxy data, and any such estimate imposes structured, data-dependent distortions whose direction depends on which corpus happened to be chosen.
The identity avoids this: ignorance may be preferable to bias.

\subsection{The Covariance of Large Models Approaches the Identity}
\label{sec:method:idcov:theory}

Further, the isotropy assumption is a theoretically well-grounded approximation of the true covariance for large models.
The Hessian of deep networks exhibits a characteristic spectral pattern---a small number of large eigenvalues with a vast bulk near zero \citep{sagun-2017-eigenvalueshessian, pennington-2018-spectrumfisher, karakida-2019-universalstatistics}---and inverting it amplifies the near-zero eigenvalues, so the damping term $\lambda I$ universally added for stabilization dominates in most parameter directions, causing the damped inverse $(F + \lambda I)^{-1}$ to converge to approximately $(1/\lambda) I$.
\citet{li-2025-influencefunctions} verify empirically that for LLMs, the damping overwhelms the Hessian so the damped inverse is effectively proportional to the identity, and \citet{kwon-2023-datainfefficiently} independently observe that iterative Hessian inversion collapses to the identity baseline on a 13B-parameter model.
Weight decay imposes an isotropic Gaussian prior \citep{bishop-2006-patternrecognition}, and pretrained language models have very low intrinsic dimensionality relative to their parameter count \citep{aghajanyan-2021-intrinsicdimensionality, hu-2021-loralowrank}, so the posterior is driven by this isotropic prior in all but a small subspace.
The isotropy assumption has also been employed, often implicitly, across data attribution \citep{pruthi-2020-estimatingtraining, charpiat-2019-inputsimilarity, yang-2024-revisitextend, jaburi-2025-mitigatingemergent, kowal-2026-conceptinfluence}, out-of-distribution detection \citep{bergamin-2022-modelagnosticoutofdistribution, zhdanov-2025-identitycurvature}, and dataset pruning \citep{paul-2021-deeplearning}, consistently matching or outperforming more elaborate curvature corrections.
A detailed review of these theoretical results and empirical precedents is provided in \cref{app:theory-empirical}.

\subsection{Estimating Aleatoric Uncertainty}
\label{sec:method:aleatoric}

The same Taylor expansion as in \cref{eq:taylor} can be used to derive an estimate of aleatoric uncertainty.
Applying it to the Bernoulli variance $h(\theta) = p(y_c \mid x, \theta)(1 - p(y_c \mid x, \theta))$ and taking expectations gives:
\begin{align}
\mathbb{E}_{\theta}\!\left[h(\theta)\right]
&\approx
h(\theta^*)
+ \left.\nabla_\theta h(\theta)\right|_{\theta^*}^{\top}
\left(\mathbb{E}_{\theta}[\theta] - \theta^*\right)
\end{align}
The first-order term $\nabla_\theta h(\theta^*)^\top (\mathbb{E}_\theta[\theta] - \theta^*)$ vanishes: combining the Taylor approximation with the assumption $p(y_c \mid x, \theta^*) = \mathbb{E}_\theta[p(y_c \mid x, \theta)]$ required for the variance decomposition (\cref{sec:bg}) yields $\mathbb{E}_\theta[\theta] = \theta^*$ (proof in \cref{app:aleatoric-proof}).
Thus, the estimate of aleatoric uncertainty reduces to $h(\theta^*) = p(y_c \mid x, \theta^*)(1 - p(y_c \mid x, \theta^*))$.
Beyond the Taylor expansion, the only additional requirement is $p(y_c \mid x, \theta^*) = \mathbb{E}_\theta[p(y_c \mid x, \theta)]$, which is itself necessary for the uncertainty decomposition.

\subsection{Extension to Sequences}
\label{sec:method:sequences}

The derivations above treat $y_c$ as a single discrete output.
For generative language models, the object of interest is a sequence $y = (y_{c_1}, \ldots, y_{c_T})$.
A direct extension would consider $\mathrm{Var}_\theta[p(y \mid x, \theta)]$, but the joint probability scales exponentially with sequence length, making it unsuitable for comparing sequences of different lengths.
Instead, we apply the Taylor expansion to the mean predicted probability,
\begin{equation}
\bar{p}(y \mid x, \theta) = \frac{1}{T}\sum_{t=1}^T p(y_{c_t} \mid y_{<t}, x, \theta),
\end{equation}
preserving consistency with the single-token derivation in \cref{sec:method}.
This yields
\begin{equation}
\mathrm{Var}_\theta[\bar{p}(y \mid x, \theta)] \approx \bar{g}^\top \mathrm{Cov}[\theta]\, \bar{g},
\end{equation}
where $\bar{g} = \frac{1}{T} \sum_{t=1}^T \nabla_\theta p(y_{c_t} \mid y_{<t}, x, \theta^*)$.
Under isotropic covariance, epistemic uncertainty reduces to $\|\bar{g}\|^2$, computable from a single backward pass.

Expanding $\|\bar{g}\|^2 = \bigl\|\frac{1}{T}\sum_t g_t\bigr\|^2$ yields $\frac{1}{T^2}\bigl(\sum_t \|g_t\|^2 + \sum_{t \neq s} g_t^\top g_s\bigr)$.
The cross-terms $g_t^\top g_s$ capture correlations in parameter space between token predictions, present in the sequence-level formulation but absent from an average of per-token norms.
These cross-terms allow the sequence-level estimate to reflect uncertainty about the sequence as a whole, rather than treating each token independently.
If these correlations are negligible, the two approaches coincide; if significant, the sequence-level estimate captures them at no additional cost, while per-token backward passes scale linearly with $T$.

Aleatoric uncertainty extends symmetrically as the mean per-token Bernoulli variance, $\frac{1}{T}\sum_{t=1}^T p(y_{c_t} \mid y_{<t}, x, \theta^*)(1 - p(y_{c_t} \mid y_{<t}, x, \theta^*))$, requiring only the forward pass.

\section{Experiments}
\label{sec:experiments}

We first validate $\|g\|^2$ against MCMC estimates on synthetic problems (\cref{sec:experiments:validation}), then investigate the utility of aleatoric and epistemic uncertainty for predicting answer correctness in LLM question answering (\cref{sec:experiments:downstream}).

\subsection{Validation}
\label{sec:experiments:validation}

\begin{table}[t]
    \centering
    \small
    \begin{subtable}[t]{0.33\textwidth}
        \centering
        \begin{tabular}{lccc}
            \toprule
            & Linear & XOR & Rings \\
            \midrule
            \multicolumn{4}{l}{\textit{Epistemic (GN)}} \\
            $r$    & 0.95 & 0.65 & 0.86 \\
            $\rho$ & 0.99 & 0.68 & 0.44 \\
            \multicolumn{4}{l}{\textit{Epistemic (LA)}} \\
            $r$    & 0.95 & 0.68 & 0.86 \\
            $\rho$ & 0.99 & 0.70 & 0.46 \\
            \multicolumn{4}{l}{\textit{Aleatoric}} \\
            $r$    & 0.99 & 0.76 & 0.95 \\
            $\rho$ & 1.00 & 0.74 & 0.58 \\
            \bottomrule
        \end{tabular}
        \caption{Binary classification}
        \label{tab:validation-binary}
    \end{subtable}\hfill
    \begin{subtable}[t]{0.33\textwidth}
        \centering
        \begin{tabular}{lccc}
            \toprule
            & Clusters & Spirals & Rings \\
            \midrule
            \multicolumn{4}{l}{\textit{Epistemic (GN)}} \\
            $r$    & 0.86 & 0.76 & 0.88 \\
            $\rho$ & 0.97 & 0.91 & 0.97 \\
            \multicolumn{4}{l}{\textit{Aleatoric}} \\
            $r$    & 0.95 & 0.96 & 0.96 \\
            $\rho$ & 0.99 & 0.97 & 0.98 \\
            \bottomrule
        \end{tabular}
        \caption{Multiclass classification}
        \label{tab:validation-multiclass}
    \end{subtable}\hfill
    \begin{subtable}[t]{0.27\textwidth}
        \centering
        \begin{tabular}{lcc}
            \toprule
            & Linear & Nonlin.\ \\
            \midrule
            \multicolumn{3}{l}{\textit{Epistemic (GN)}} \\
            $r$    & 0.98 & 0.73 \\
            $\rho$ & 0.99 & 0.81 \\
            \multicolumn{3}{l}{\textit{Epistemic (LA)}} \\
            $r$    & 1.00 & 0.93 \\
            $\rho$ & 1.00 & 0.97 \\
            \bottomrule
        \end{tabular}
        \caption{Regression}
        \label{tab:validation-regression}
    \end{subtable}
    \caption{Pearson ($r$) and Spearman ($\rho$) correlations between our estimates and MCMC estimates. GN: gradient norm $\|g\|^2$; LA: Laplace $g^\top H^{-1} g$. Aleatoric: $p(y_c \mid x, \theta^*)(1 - p(y_c \mid x, \theta^*))$.}
    \label{tab:validation}
\end{table}

We compare our estimates directly against the quantity they are designed to approximate, rather than using out-of-distribution (OOD) detection as a proxy.
OOD detection assumes that inputs far from the training data produce high epistemic uncertainty, but Bayesian epistemic uncertainty depends on the space of plausible parameterizations, not on distance from training data alone---a linear classifier, for instance, cannot exhibit high epistemic uncertainty far from its boundary regardless of how distant the input is from any training point.
This disconnect has been observed in practice \citep{ulmer-2020-trustissues}, so failures on OOD benchmarks may reflect a mismatch between the validation assumption and the quantity being measured rather than a deficiency of the method.
On synthetic problems where the parameter count permits full posterior inference, we use Hamiltonian Monte Carlo (HMC) \citep{betancourt-2018-conceptualintroduction} with dual-averaging step-size adaptation \citep{nesterov-2009-primaldualsubgradient, hoffman-2014-nouturnsampler} to compute $\operatorname{Var}_\theta[p(y_c \mid x, \theta)]$ and measure how well $\|g\|^2$ tracks it, using Pearson and Spearman correlation.
We also evaluate the Laplace approximation $g^\top H^{-1} g$ as a point of comparison.

We conduct experiments across classification (logistic/softmax regression and multilayer perceptrons (MLPs) on 2D problems of varying complexity), regression (single-hidden-layer MLP on 1D problems), and a scaling study (12 to ${\sim}10^6$ parameters).
Full setup details are given in \cref{app:validation-details}.

\subsubsection{Classification and Regression}

\begin{figure}[h]
    \centering
    \begin{subfigure}[t]{0.235\textwidth}
        \centering
        \includegraphics[width=\textwidth]{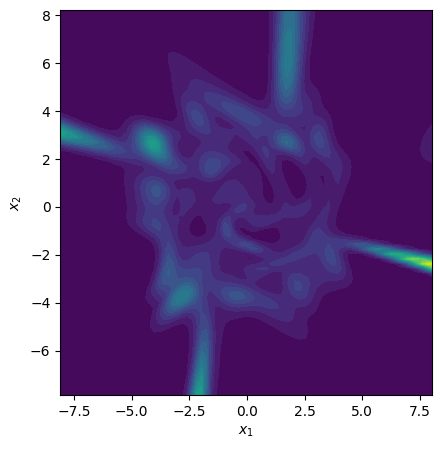}
        \caption{Epistemic, MCMC}
    \end{subfigure}\hfill
    \begin{subfigure}[t]{0.235\textwidth}
        \centering
        \includegraphics[width=\textwidth]{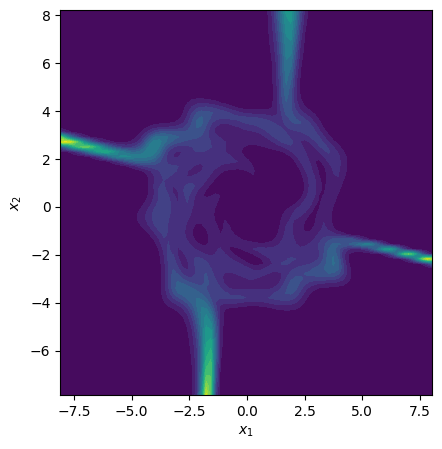}
        \caption{Epistemic, GN}
    \end{subfigure}\hfill
    \begin{subfigure}[t]{0.235\textwidth}
        \centering
        \includegraphics[width=\textwidth]{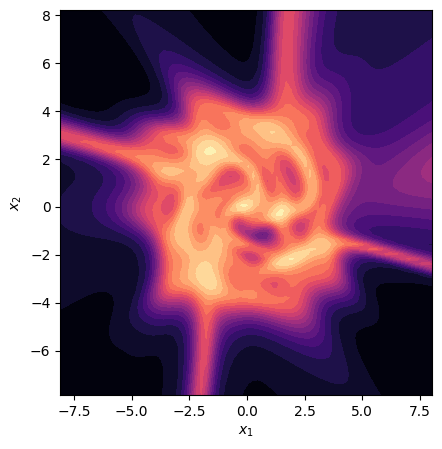}
        \caption{Aleatoric, MCMC}
    \end{subfigure}\hfill
    \begin{subfigure}[t]{0.235\textwidth}
        \centering
        \includegraphics[width=\textwidth]{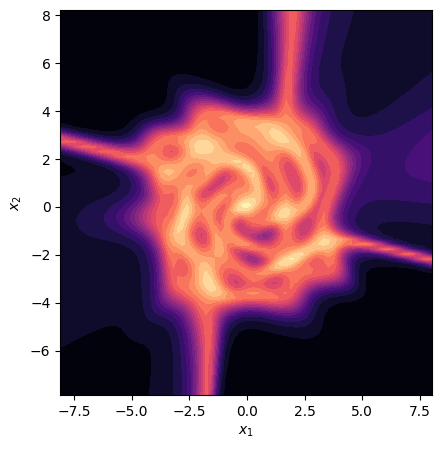}
        \caption{Aleatoric, point est.}
    \end{subfigure}
    \caption{Multiclass spirals uncertainty maps. Left two panels: epistemic uncertainty (MCMC vs.\ gradient norm $\|g\|^2$). Right two panels: aleatoric uncertainty (MCMC vs.\ point estimate). All maps are individually normalized to $[0, 1]$. Additional problems in \cref{app:classification-additional}.}
    \label{fig:classification}
\end{figure}

Across six classification problems, $\|g\|^2$ consistently tracks the MCMC estimates (\cref{tab:validation}).
Correlations are highest when the model is correctly specified and lower for MLPs with anisotropic posteriors, though the correct spatial pattern is always recovered (\cref{fig:classification}).
The Laplace approximation provides almost no improvement over $\|g\|^2$ in classification, indicating that posterior curvature is negligible in this setting.
Aleatoric point estimates are uniformly strong across all multiclass problems.

In regression, the isotropy assumption breaks down: on a nonlinear problem the Laplace approximation substantially outperforms $\|g\|^2$ (\cref{tab:validation,app:regression-figure}), confirming that the isotropy assumption rather than the shared Taylor expansion is the primary source of error.

\subsubsection{Scaling with Model Size}

\begin{figure}[h]
    \centering
    \begin{subfigure}[t]{0.48\textwidth}
        \centering
        \includegraphics[width=\textwidth]{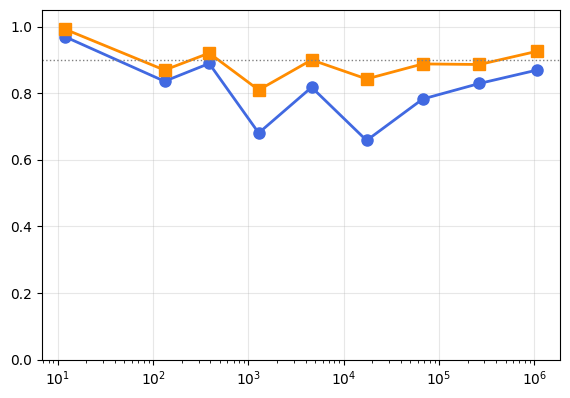}
        \caption{Spearman correlation}
    \end{subfigure}
    \hfill
    \begin{subfigure}[t]{0.48\textwidth}
        \centering
        \includegraphics[width=\textwidth]{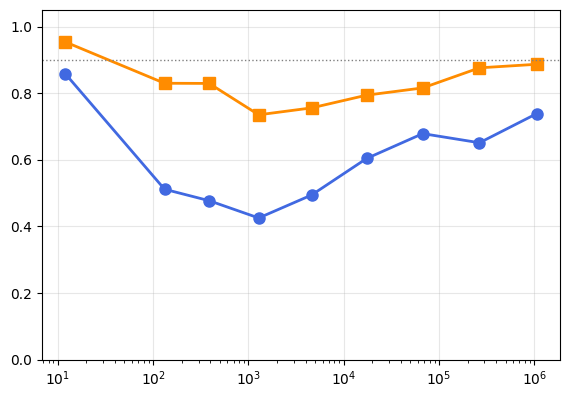}
        \caption{Pearson correlation}
    \end{subfigure}
    \caption{Correlation between our estimates and MCMC estimates as a function of model size (number of parameters) on a concentric rings problem. Both epistemic (blue) and aleatoric (orange) correlations follow a U-shaped trajectory, dipping at intermediate scales and recovering at larger model sizes. Full per-model results in \cref{tab:scaling-full}.}
    \label{fig:scaling}
\end{figure}

The theoretical arguments in \cref{sec:method:idcov:theory} predict that the identity becomes a better proxy as network scale increases.
We test this by training models from 12 to approximately $10^6$ parameters on a concentric rings problem.
As \cref{fig:scaling} shows, the epistemic correlation follows a U-shaped trajectory: high for the smallest models (where the posterior is trivially isotropic), dipping at intermediate scales (where the posterior is anisotropic but spectral concentration has not yet taken effect), and recovering to $\rho = 0.87$ at approximately $10^6$ parameters.
The recovery begins roughly when the number of parameters exceeds the training samples, placing the model in the overparameterized regime where the FIM becomes increasingly low-rank.

We additionally extend the scaling study using mean-field variational inference as the reference, which is tractable at substantially larger scales than HMC, sweeping from $\sim$$10^2$ to $\sim$$10^8$ parameters (\cref{app:scaling-vi}).
The same U-shape appears in Pearson $r$ with the minimum at $|\text{parameters}| \approx |\text{training samples}|$, and Spearman $\rho$ stays above $0.96$ throughout.
That the same pattern appears with two independent reference methods suggests it reflects a property of the approximation rather than an artifact of either reference.
The dip falling at the interpolation threshold suggests the approximation is weakest precisely in the regime where these synthetic experiments operate, and should improve at the scales of LLMs.

\subsection{Utility of Uncertainty for Question Answering}
\label{sec:experiments:downstream}

We now investigate when aleatoric and epistemic uncertainty carry useful signal for predicting answer correctness in LLM question answering.
We evaluate four LLMs on TriviaQA \citep{joshi-2017-triviaqalarge} and TruthfulQA \citep{lin-2022-truthfulqameasuring}, measuring each uncertainty type's ability to predict correctness via AUROC.
Following \citet{farquhar-2024-detectinghallucinations}, we use their semantic equivalence criterion and compare against na\"ive and semantic entropy \citep{kuhn-2022-semanticuncertainty} as well as P(True), which prompts the model to assess its own answer correctness \citep{kadavath-2022-languagemodels}.
Several other methods are inapplicable to our setting: deep ensembles require training multiple models, the Laplace approximation faces the proxy-data concerns of \cref{sec:method:idcov:proxy}, MC dropout \citep{gal-2016-dropoutbayesian} requires the model to have been trained with dropout (which the LLMs we evaluate are not), and Deep Evidential Regression \citep{amini-2020-deepevidential} requires retraining with a modified loss.
Results are averaged over four models and 300 bootstrap runs; full details in \cref{app:qa-details}.

\begin{wraptable}{r}{0.50\textwidth}
    \centering
    \small
    \begin{tabular}{lcc}
        \toprule
        Method & TriviaQA & TruthfulQA \\
        \midrule
        P(True)          & $\mathbf{0.69} \pm 0.06$ & $0.55 \pm 0.06$ \\
        Sem.\ Entropy    & $0.55 \pm 0.03$          & $0.54 \pm 0.08$ \\
        Na\"ive Entropy  & $0.52 \pm 0.04$          & $0.51 \pm 0.08$ \\
        \midrule
        Aleatoric        & $0.60 \pm 0.04$          & $0.60 \pm 0.09$ \\
        Epistemic        & $0.52 \pm 0.07$          & $0.55 \pm 0.07$ \\
        Epi.\ \& Alea.\   & $0.61 \pm 0.05$          & $\mathbf{0.63} \pm 0.08$ \\
        \bottomrule
    \end{tabular}
    \caption{AUROC (mean $\pm$ std over 300 bootstrap runs, 4 LLMs) for predicting answer correctness. Higher is better; $0.50$ is chance. Best per column in bold.}
    \label{tab:qa-results}
\end{wraptable}

\subsubsection{Results}

\cref{tab:qa-results} reports the AUROC scores averaged across models.
On TriviaQA, P(True) ($0.69$) dominates.
On TruthfulQA, the pattern reverses: the combined estimate achieves $0.63$, the highest score on this benchmark, significantly outperforming P(True) ($0.55$) and the entropy baselines ($p < 0.01$ after Benjamini--Hochberg (BH) correction; \cref{app:qa-significance}).

This divergence between the two benchmarks is the most instructive finding.
TriviaQA tests factual recall, where a model may be equally confident in correct and incorrect answers, so uncertainty and correctness are largely independent.
TruthfulQA targets common misconceptions with genuinely ambiguous answer spaces, creating both inherent output ambiguity and epistemic conflict between popular and truthful answers.
In this setting, the aleatoric estimate ($0.60$) reflects output-level hedging, the epistemic estimate ($0.55$) captures parameter-level sensitivity, and their combination ($0.63$) outperforms all baselines.
P(True) loses its advantage because the model's self-assessment is precisely what TruthfulQA is designed to defeat, while aleatoric and epistemic uncertainty carry genuinely useful signal, suggesting that these uncertainty types are most informative when the task involves conflict between plausible parameterizations rather than factual memorization.
The epistemic estimate and P(True) are only weakly correlated (Spearman $\rho \approx {-}0.2$ on both benchmarks; \cref{app:qa-correlation}), confirming they capture largely distinct signal.

The per-model breakdown (\cref{tab:qa-per-model} in \cref{app:qa-per-model}) reveals substantial model-level variation, but several trends are consistent.
The benchmark divergence is universal: on TruthfulQA, the combined estimate is at least on par with the best baseline for every model, while on TriviaQA the reverse holds for three of four models.
The relative utility of aleatoric and epistemic uncertainty is model-dependent: both Llama models favor aleatoric uncertainty on TruthfulQA, while OLMo and Phi-4 favor epistemic.
Models from the same family behaving alike suggests training data as a driver---models that have seen more relevant data would have less epistemic uncertainty to exploit, making output ambiguity the dominant signal, while models with genuine knowledge gaps benefit more from the epistemic estimate.
Notably, the aleatoric estimate alone ($0.60$) is competitive; the epistemic component provides a significant lift when combined ($0.60 \to 0.63$, paired bootstrap $p = 0.018$ on TruthfulQA, not significant on TriviaQA; \cref{app:qa-significance}) and on individual models carries substantial independent signal (e.g., Phi-4 epistemic alone: $0.63$).

\subsubsection{Computational Cost}

Beyond accuracy, the gradient-based estimates offer a substantial computational advantage.
The entropy-based methods sample $K$ alternative completions to estimate predictive entropy, and P(True) similarly samples $K$ alternative completions before evaluating its own answer against them; our method requires only a single backward pass after generation and computes both epistemic ($\|\bar g\|^2$) and aleatoric ($\bar p (1 - \bar p)$) estimates from the same pass (full per-sample pass counts in \cref{app:complexity}).
We benchmark each method on TruthfulQA on a single NVIDIA H100 GPU, using default settings from each method's respective paper, reporting the mean and standard deviation of per-sample wall-clock time (excluding the shared generation step; full per-model results in \cref{tab:timing-full}).
The gradient norm yields a $46$--$107\times$ speedup over the baselines.

This advantage matters most precisely in the regime where uncertainty quantification is most needed: large-scale generation, where the cost of running multiple sampling passes per query is prohibitive.

\section{Conclusion}
\label{sec:conclusion}

By approximating uncertainty via a first-order Taylor expansion under isotropic covariance, we reduce epistemic uncertainty to the squared norm of the prediction gradient and aleatoric uncertainty to the Bernoulli variance of the point prediction, giving a complete uncertainty decomposition from a single forward-backward pass through an unmodified pretrained model.

Validation against reference MCMC estimates on synthetic problems shows strong correspondence in classification (Spearman $\rho$ of $0.44$--$0.99$ across settings), with an improving trend at larger model sizes that supports the isotropy assumption.
The downstream question answering experiments reveal that uncertainty estimates are most informative when the model faces genuine conflict between plausible parameterizations (as on TruthfulQA, where at least one uncertainty estimate exceeds all baselines for every model), rather than when correctness depends on factual memorization, though the relative utility of aleatoric and epistemic uncertainty varies substantially between models.
More broadly, the near-chance epistemic AUROC on TriviaQA suggests that epistemic uncertainty may not be as useful for hallucination detection as previously assumed \citep{xiao-2021-hallucinationpredictive, han-2025-negativenudging, park-2026-efficientepistemic, liu-2026-enhancinghallucination}, since factual errors need not coincide with parameter-level disagreement; gradient-based uncertainty captures a complementary signal to self-assessment methods like P(True), with the two excelling on fundamentally different question types.
More generally, even when the Bayesian calibration of the squared gradient norm is approximate, it retains a meaningful ranking of inputs as a measure of local sensitivity to parameter perturbations.

\paragraph{Limitations.}
The estimates are on the scale of squared gradient norms, which lack intuitive interpretation.
The unknown global scale $\sigma^2$ from $\operatorname{Cov}[\theta] = \sigma^2 I$ cancels in every ranking-based comparison within a single model (AUROC, Pearson, Spearman), so the choice of $\sigma^2 = 1$ is without loss of generality for our experiments; but it does not cancel in two settings where it would otherwise be useful: (a) the relative scaling between the aleatoric and epistemic uncertainty estimates, so their ratio is not meaningful in absolute terms, and (b) cross-model comparison, where each model has its own unknown $\sigma^2$.
On the latter, training an answer-correctness classifier on the uncertainty estimates from three models and evaluating on the fourth yields chance-or-below performance, with the relationship between gradient norm and correctness occasionally inverting on the held-out model, even after normalizing by the squared parameter norm (\cref{app:qa-lomo}).
Calibrating $\sigma^2$ without downstream supervision---which would address both---is an open problem and a natural direction for future work.
The isotropy assumption itself, while well-motivated at scale, introduces measurable error at intermediate model sizes and in regression settings where the posterior is highly anisotropic.
The scaling study validates up to ${\sim}10^8$ parameters (with mean-field VI as the reference; HMC is tractable up to ${\sim}10^6$) while the LLM experiments operate at $10^9$--$10^{10}$; although the trend is monotonically improving and multiple lines of evidence support continued improvement, there is no formal guarantee.
On the downstream task, the substantial variance across models and bootstrap runs currently precludes reliable deployment for assessing individual predictions.

\section*{Acknowledgements}

This work was supported by the Danish Data Science Academy, which is funded by the Novo Nordisk Foundation (NNF21SA0069429) and VILLUM FONDEN (40516).

\bibliography{references}

\appendix

\section{Proxy Covariance Bias Experiment}
\label{app:hessian-bias}

This appendix provides the full experimental details for the synthetic proxy bias experiment summarized in \cref{sec:method:idcov:proxy}.

\paragraph{Setup.}
We train a binary classifier---a two-hidden-layer MLP with $\tanh$ activations and 1\,185 parameters---on three 2D problems: a linearly separable boundary, an XOR pattern, and concentric rings.
Each problem is designed so that the decision boundary is spatially symmetric.
For each problem, we split the training data into two halves by spatial location (top vs.\ bottom) and compute the empirical Fisher information matrix (FIM) from each half separately, obtaining two proxy Hessians $H_A$ (from top-half data) and $H_B$ (from bottom-half data).
We then evaluate epistemic uncertainty under three covariance assumptions: the identity ($C = I$, yielding $\|g\|^2$), the Laplace approximation using $H_A^{-1}$, and the Laplace approximation using $H_B^{-1}$, writing $U_A(x) = g^\top H_A^{-1} g$ and $U_B(x) = g^\top H_B^{-1} g$ for the two proxy estimates.
All uncertainty maps are evaluated on a dense grid covering the input space and individually normalized to $[0, 1]$ for comparison.

\paragraph{Results.}
On all three problems, the identity produces uncertainty estimates that respect the spatial symmetry of the decision boundary, while each proxy Hessian introduces a severe asymmetry: it suppresses uncertainty in the half of input space where its data is concentrated and inflates it in the complementary half.
\cref{tab:hessian-bias} reports the full results.

\begin{table}[h!]
    \centering
    \small
    \begin{tabular}{lccccccc}
        \toprule
        & \multicolumn{3}{c}{Top/bottom ratio} & \multicolumn{3}{c}{Welch $t$-test on $\log(U_A/U_B)$} \\
        \cmidrule(lr){2-4} \cmidrule(lr){5-7}
        Problem & $r_{\mathrm{id}}$ & $r_A$ & $r_B$ & $t$ & $p$ & Cohen's $d$ \\
        \midrule
        Linear & 0.93 & 0.21 & 4.56 & $-8.1$ & $3.0 \times 10^{-14}$ & $-1.0$ \\
        XOR    & 1.45 & 0.62 & 4.39 & $-19.6$ & $3.8 \times 10^{-50}$ & $-2.5$ \\
        Rings  & 1.06 & 0.05 & 19.10 & $-22.2$ & $2.0 \times 10^{-61}$ & $-2.8$ \\
        \bottomrule
    \end{tabular}
    \caption{Proxy covariance bias across three synthetic problems. $r_{\mathrm{id}}$, $r_A$, $r_B$: ratio of mean top-half to mean bottom-half normalized uncertainty under the identity, proxy $H_A^{-1}$ (top-half data), and proxy $H_B^{-1}$ (bottom-half data), respectively. A ratio of $1.0$ indicates perfect spatial symmetry. The Welch $t$-test is performed on the log-ratio map $\log(U_A(x)/U_B(x))$ between the two halves of the input space.}
    \label{tab:hessian-bias}
\end{table}

The symmetry ratio under isotropic covariance, $r_{\mathrm{id}}$ (ratio of mean uncertainty in each half of the input space), remains close to $1.0$ on all three problems, confirming that it preserves the true spatial symmetry.
The corresponding ratios under each proxy covariance, $r_A$ and $r_B$ (same metric, using $H_A^{-1}$ and $H_B^{-1}$ respectively), deviate strongly in opposite directions, with the effect increasing with decision boundary complexity: from the linear problem ($d = -1.0$) through XOR ($d = -2.5$) to rings ($d = -2.8$).
All effects are highly significant ($p < 10^{-13}$).
The weaker effect on the linear problem is expected: a linear decision boundary constrains fewer parameter directions, so there is less room for the proxy to distort the covariance structure.

\subsection{Language Model Extension}
\label{app:hessian-bias-bert}

This appendix provides the experimental details for the multiclass DistilBERT proxy bias experiment summarized in \cref{sec:method:idcov:proxy}.

We fine-tune a single 20-class DistilBERT on all of 20~Newsgroups (3 epochs, AdamW with $\eta = 2 \times 10^{-5}$, weight decay $10^{-2}$, $79.5\%$ test accuracy).
We construct proxy Hessians from four topical subsets of the data---\texttt{sci.*} (science), \texttt{comp.*} (computers), \texttt{rec.*} (recreation), and \texttt{talk.*} (politics/religion)---together with all $10$ non-empty pair and triple combinations of these four groups, a representative sample drawn proportionally from all four groups, and a full Hessian pooling all proxy samples together.
We apply a last-layer Laplace approximation \citep{kristiadi-2020-beingbayesian}: for each covariance, we compute the empirical FIM on the classifier head parameters ($768 \times 20 + 20 = 15{,}380$ parameters) from $200$ proxy samples (distributed equally across the included groups for multi-domain proxies), with prior precision $\lambda = 1$.
We evaluate epistemic uncertainty $g^\top C\, g$ under each covariance $C$ on a shared test set of $200$ samples drawn from all $20$ categories, using the full Hessian estimate as the reference.
The per-domain deviations and CV across domain combinations reported in \cref{tab:hessian-bias-bert-cv-main,tab:hessian-bias-bert-combos} are computed against this reference; \cref{tab:hessian-bias-bert-corr} additionally reports the overall Spearman and Pearson correlation of each single-domain proxy, the identity, and the representative sample with the full Hessian.

\begin{table}[h!]
    \centering
    \small
    \begin{tabular}{lcc}
        \toprule
        Covariance assumption & Spearman $\rho$ & Pearson $r$ \\
        \midrule
        all domains (representative) & 0.984 & 0.940 \\
        identity (ours)              & 0.979 & 0.889 \\
        \texttt{sci} only            & 0.962 & 0.830 \\
        \texttt{comp} only           & 0.928 & 0.751 \\
        \texttt{rec} only            & 0.946 & 0.833 \\
        \texttt{talk} only           & 0.920 & 0.763 \\
        \bottomrule
    \end{tabular}
    \caption{Spearman and Pearson correlation between each covariance's uncertainty estimates and those obtained with the full Hessian, on the shared test set.}
    \label{tab:hessian-bias-bert-corr}
\end{table}

\begin{table}[h!]
    \centering
    \small
    \begin{tabular}{ll}
        \toprule
        Domains in proxy & CV (sorted within group) \\
        \midrule
        identity (none)    & $0.21$ \\
        $1$ (single domain)  & $0.33, 0.40, 0.57, 0.60$ \\
        $2$ (pairs)          & $0.22, 0.31, 0.35, 0.37, 0.54, 0.61$ \\
        $3$ (triples)        & $0.17, 0.19, 0.35, 0.54$ \\
        $4$ (representative) & $0.13$ \\
        \bottomrule
    \end{tabular}
    \caption{CV across domains for proxies built from all $14$ combinations of the four topical groups, grouped by the number of domains included.}
    \label{tab:hessian-bias-bert-combos}
\end{table}

\section{Theoretical and Empirical Support for Isotropic Covariance}
\label{app:theory-empirical}

This appendix provides the detailed theoretical arguments and empirical precedents supporting the isotropy assumption summarized in \cref{sec:method:idcov:theory}.

\subsection{Hessian Structure Simplifies with Scale}

A convergent body of theoretical and empirical work shows that the Hessian of deep networks exhibits a characteristic spectral pattern: a small number of large eigenvalues with a vast bulk near zero.
Much of this theory has been developed for the FIM, a standard positive semi-definite approximation to the Hessian:
\citet{amari-2019-fisherinformation} prove that the FIM is approximately unit-wise block-diagonal, with off-block elements of order $O(1/\sqrt{n})$ for layer width $n$;
within each block, \citet{dauncey-2024-approximationsfisher} demonstrate diagonal dominance, with diagonal entries roughly five times larger than off-diagonal ones;
and the eigenvalue distribution has been characterized analytically via mean-field theory \citep{karakida-2019-universalstatistics} and nonlinear random matrix theory \citep{pennington-2018-spectrumfisher}.
\citet{karakida-2021-pathologicalspectra} show that this bulk spectral concentration holds equally for the empirical FIM and the generalized Gauss--Newton matrix.
Empirical studies of the full Hessian confirm the same two-component structure in trained networks \citep{sagun-2017-eigenvalueshessian, sagun-2018-empiricalanalysis, papyan-2020-tracesclass}.

This spectral structure has a direct consequence for the covariance.
Inverting the Hessian amplifies the near-zero eigenvalues, so the damping term $\lambda I$ universally added for stabilization dominates in most parameter directions, causing the damped inverse $(F + \lambda I)^{-1}$ to converge to approximately $(1/\lambda) I$.
This effect becomes more pronounced at scale: \citet{li-2025-influencefunctions} show empirically that for LLMs, the damping term overwhelms the Hessian, so that the damped inverse is effectively proportional to the identity.
Corroborating this, \citet{kwon-2023-datainfefficiently} report that the LiSSA algorithm for iterative inverse-Hessian approximation collapses to the Hessian-free (identity) baseline across all tasks on a 13B-parameter model, which they attribute to the high dimensionality of large-scale models.
These results also have implications for the structural error in \cref{eq:diaglaplace-decomp} and \cref{eq:identity-decomp}: the off-block-diagonal decay at $O(1/\sqrt{n})$ and the within-block diagonal dominance together suggest that $\|\Sigma_{\mathrm{off}}\|_{\mathrm{op}}$ shrinks relative to $\|\Sigma_{\mathrm{diag}}^*\|_{\mathrm{op}}$ as scale increases, so that term~(i) becomes a diminishing fraction of the total error for both the identity and any diagonal approximation.

\subsection{Precedent from Laplace Approximations}

The Laplace approximation for neural networks provides perhaps the most direct precedent for simplifying the covariance structure without sacrificing downstream performance: practitioners routinely employ drastic simplifications of the Hessian with little loss.
The full Hessian or generalized Gauss--Newton (GGN) matrix is typically replaced by a diagonal \citep{kirkpatrick-2017-overcomingcatastrophic, ritter-2018-scalablelaplace}, Kronecker-factored \citep{ritter-2018-scalablelaplace, eschenhagen-2023-kroneckerfactoredapproximate}, or block-diagonal approximation.
More aggressive still, last-layer Laplace restricts the posterior to only the final layer's parameters \citep{kristiadi-2020-beingbayesian, daxberger-2021-laplaceredux}, and subnetwork Laplace \citep{daxberger-2021-bayesiandeep} selects an arbitrary subset of parameters for Bayesian treatment.
\citet{daxberger-2021-laplaceredux} systematically compare these approximations and find that the cheapest variants---diagonal and last-layer---often match or exceed the predictive performance of more faithful Hessian estimates, suggesting that the precise structure of the covariance matters far less than one might expect.
Our isotropy assumption goes further by replacing per-parameter curvature magnitudes with a uniform scalar, but the spectral results above imply that at scale the damped inverse is already approximately proportional to the identity, and obtaining accurate magnitudes without the true training data introduces structured distortions (\cref{sec:method:idcov:proxy}) that may outweigh the benefit.

\subsection{Bayesian and Optimization Arguments}

Most modern LLMs are trained with weight decay, which is equivalent to imposing a Gaussian prior with covariance proportional to the identity \citep{bishop-2006-patternrecognition, goodfellow-2016-deeplearning}.
Pretrained language models have very low intrinsic dimensionality relative to their full parameter count \citep{aghajanyan-2021-intrinsicdimensionality, hu-2021-loralowrank}, so the training data determines the posterior in only a small subspace; in the remaining directions, the posterior is driven by the isotropic prior.
Moreover, \citet{farquhar-2020-libertydepth} prove that diagonal weight-space covariance in deep networks can induce function-space distributions comparable to structured covariance approximations in shallower networks, suggesting that the off-diagonal structure our approximation discards is largely redundant.
A complementary argument comes from optimization: \citet{smith-2020-originimplicit} show that stochastic gradient descent (SGD) implicitly regularizes the squared norms of per-sample loss gradients, suppressing $\|g\|^2$ in well-learned regions while leaving it unconstrained for unfamiliar inputs, directionally consistent with the contrast needed for epistemic uncertainty estimation without an explicit covariance correction.

\subsection{Empirical Success of the Isotropy Assumption}

The isotropy assumption, whether explicit or implicit, has been employed across multiple tasks with strong empirical performance.

In out-of-distribution detection, \citet{bergamin-2022-modelagnosticoutofdistribution} propose a model-agnostic method that scores anomalies using gradient information weighted by different approximations to the Hessian; the identity performs competitively with more elaborate curvature approximations.
\citet{zhdanov-2025-identitycurvature} introduce the Identity Curvature Laplace Approximation (ICLA), which replaces the Hessian entirely with the identity and outperforms standard last-layer Laplace using the empirical FIM, GGN, and K-FAC on OOD detection benchmarks.

In data attribution, several methods define training sample influence via the gradient dot product $\nabla_\theta \ell(z_{\text{test}})^\top \nabla_\theta \ell(z_{\text{train}})$, corresponding to the influence function with the inverse Hessian replaced by the identity.
\citet{charpiat-2019-inputsimilarity} define input similarity as the cosine similarity of parameter gradients;
\citet{pruthi-2020-estimatingtraining} formalize this as TracIn;
\citet{yang-2024-revisitextend} show that this identity approximation is order-consistent with true influence in many practical regimes and that the inverse Hessian can introduce errors making it worse than the identity;
\citet{jaburi-2025-mitigatingemergent} find essentially no performance loss from using the identity for mitigating emergent behaviors in LLMs;
and \citet{kowal-2026-conceptinfluence} show that an even more aggressive double-identity approximation to concept-based influence functions matches or exceeds full EK-FAC performance at 7B scale while being 20$\times$ faster.

In dataset pruning, \citet{paul-2021-deeplearning} introduce the Gradient Norm (GraNd) score, the expected $L^2$ norm of the per-sample loss gradient, and use it to prune significant fractions of training data without sacrificing test accuracy.

\section{Aleatoric Estimates under the First-Order Taylor Approximation}
\label{app:aleatoric-proof}

From the first-order Taylor expansion used in \cref{sec:method}:
\begin{align}
p(y_c \mid x, \theta)
&\approx p(y_c \mid x, \theta^*)
+ \left.\nabla_\theta p(y_c \mid x, \theta)\right|_{\theta^*}^{\top}
(\theta - \theta^*)
\end{align}
Taking expectations on both sides:
\begin{align}
\mathbb{E}_\theta\!\left[p(y_c \mid x, \theta)\right]
&\approx p(y_c \mid x, \theta^*)
+ \left.\nabla_\theta p(y_c \mid x, \theta)\right|_{\theta^*}^{\!\top}
\left(\mathbb{E}_\theta[\theta] - \theta^*\right)
\end{align}
By the assumption that $p(y_c \mid x, \theta^*) = \mathbb{E}_\theta\!\left[p(y_c \mid x, \theta)\right]$, which is necessary for the variance-based uncertainty decomposition (\cref{sec:bg}), the left-hand side equals the first term on the right, so:
\begin{align}
\left.\nabla_\theta p(y_c \mid x, \theta)\right|_{\theta^*}^{\!\top}
\left(\mathbb{E}_\theta[\theta] - \theta^*\right)
&= 0, \\
\mathbb{E}_\theta[\theta] &= \theta^*.
\end{align}
Since $\mathbb{E}_\theta[\theta]$ and $\theta^*$ are global quantities independent of $x$, the difference $\mathbb{E}_\theta[\theta] - \theta^*$ is a single fixed vector; a single input $x$ with nonzero gradient therefore suffices to constrain it via $g^\top(\mathbb{E}_\theta[\theta] - \theta^*) = 0$.
That is, the conclusion holds exactly under the first-order Taylor approximation; the only source of error is the approximation itself.

\section{Additional Classification Results}
\label{app:classification-additional}

\begin{figure}[h!]
    \centering
    \begin{subfigure}[t]{0.235\textwidth}
        \centering
        \includegraphics[width=\textwidth]{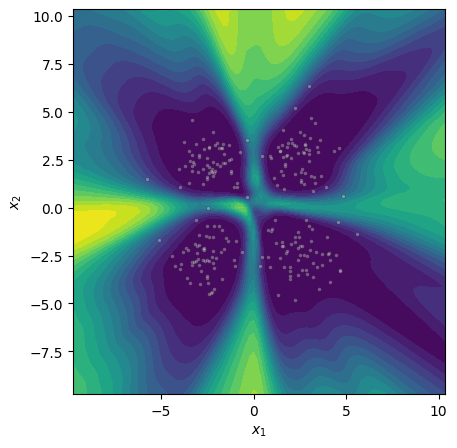}
        \caption{Epistemic, MCMC}
    \end{subfigure}\hfill
    \begin{subfigure}[t]{0.235\textwidth}
        \centering
        \includegraphics[width=\textwidth]{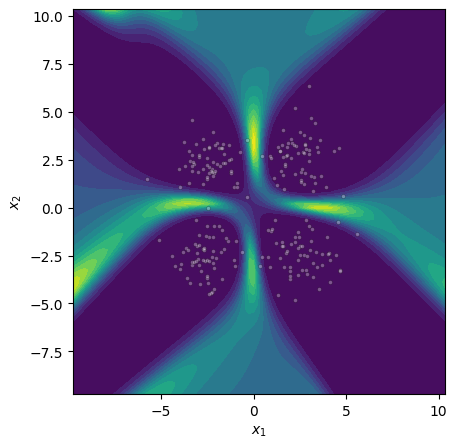}
        \caption{Epistemic, GN}
    \end{subfigure}\hfill
    \begin{subfigure}[t]{0.235\textwidth}
        \centering
        \includegraphics[width=\textwidth]{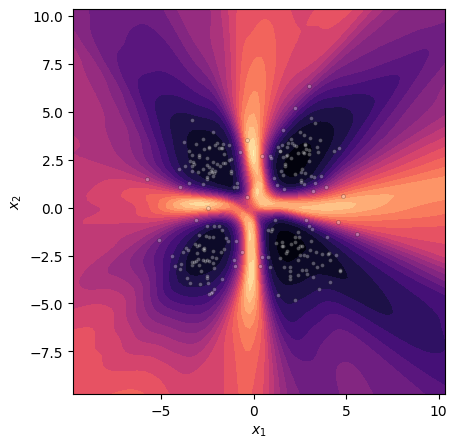}
        \caption{Aleatoric, MCMC}
    \end{subfigure}\hfill
    \begin{subfigure}[t]{0.235\textwidth}
        \centering
        \includegraphics[width=\textwidth]{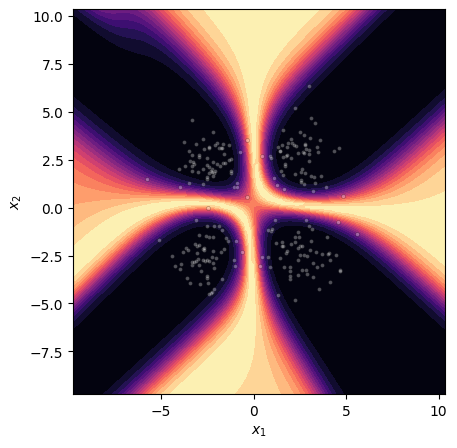}
        \caption{Aleatoric, point est.}
    \end{subfigure}
    \caption{Binary XOR uncertainty maps. Left two panels: epistemic uncertainty (MCMC vs.\ $\|g\|^2$). Right two panels: aleatoric uncertainty (MCMC vs.\ point estimate). All maps are individually normalized to $[0, 1]$.}
    \label{fig:xor-classification}
\end{figure}

\begin{figure}[h!]
    \centering
    \begin{subfigure}[t]{0.235\textwidth}
        \centering
        \includegraphics[width=\textwidth]{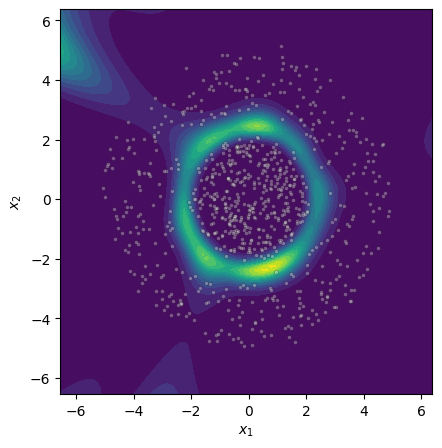}
        \caption{Binary, MCMC}
    \end{subfigure}\hfill
    \begin{subfigure}[t]{0.235\textwidth}
        \centering
        \includegraphics[width=\textwidth]{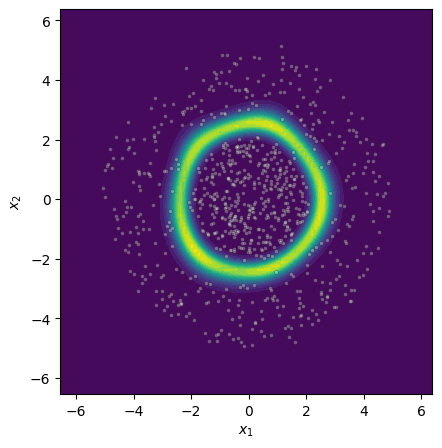}
        \caption{Binary, GN}
    \end{subfigure}\hfill
    \begin{subfigure}[t]{0.235\textwidth}
        \centering
        \includegraphics[width=\textwidth]{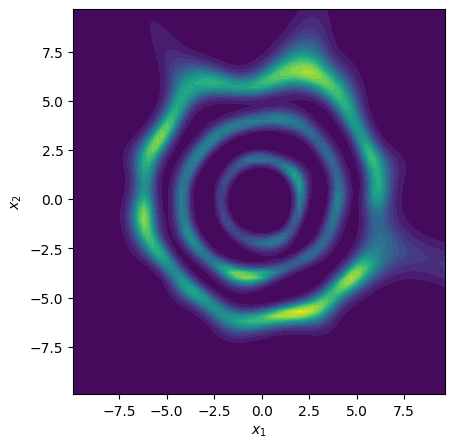}
        \caption{Multiclass, MCMC}
    \end{subfigure}\hfill
    \begin{subfigure}[t]{0.235\textwidth}
        \centering
        \includegraphics[width=\textwidth]{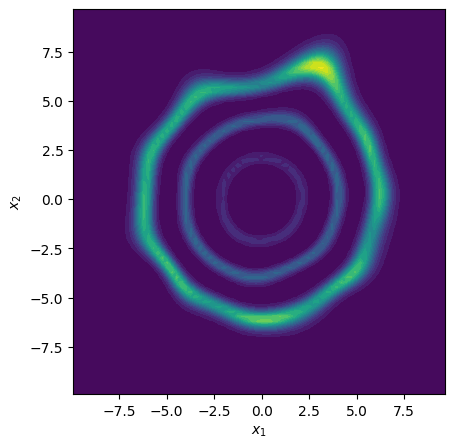}
        \caption{Multiclass, GN}
    \end{subfigure}
    \\[0.5em]
    \begin{subfigure}[t]{0.235\textwidth}
        \centering
        \includegraphics[width=\textwidth]{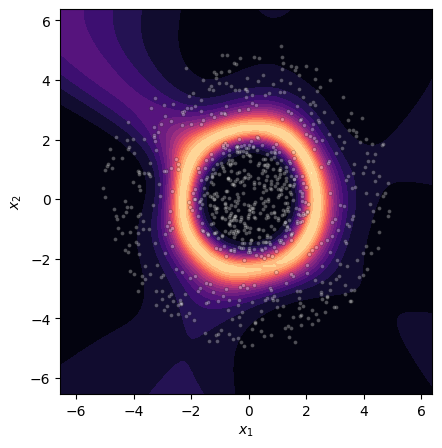}
        \caption{Binary, MCMC}
    \end{subfigure}\hfill
    \begin{subfigure}[t]{0.235\textwidth}
        \centering
        \includegraphics[width=\textwidth]{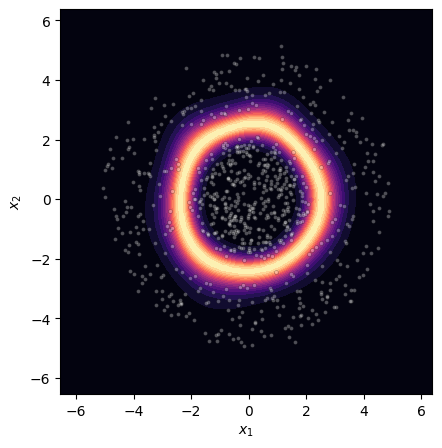}
        \caption{Binary, point est.}
    \end{subfigure}\hfill
    \begin{subfigure}[t]{0.235\textwidth}
        \centering
        \includegraphics[width=\textwidth]{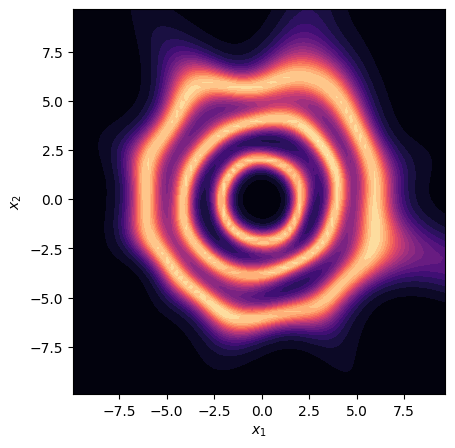}
        \caption{Multiclass, MCMC}
    \end{subfigure}\hfill
    \begin{subfigure}[t]{0.235\textwidth}
        \centering
        \includegraphics[width=\textwidth]{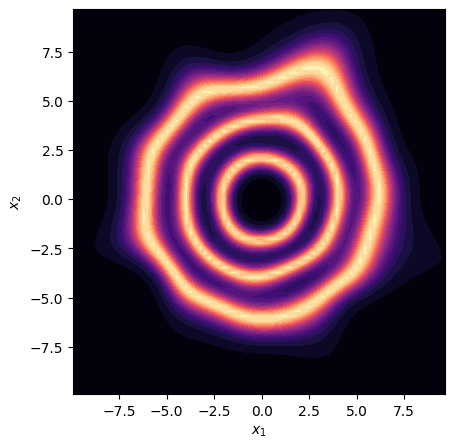}
        \caption{Multiclass, point est.}
    \end{subfigure}
    \caption{Concentric rings uncertainty maps. Top row: epistemic uncertainty (MCMC vs.\ $\|g\|^2$) for binary (left) and multiclass (right). Bottom row: aleatoric uncertainty (MCMC vs.\ point estimate). All maps are individually normalized to $[0, 1]$.}
    \label{fig:rings-classification}
\end{figure}

\cref{fig:xor-classification} shows the uncertainty maps for the binary XOR problem, and \cref{fig:rings-classification} for the concentric rings problem, both omitted from the main text for space. The binary rings problem ($\rho = 0.44$) represents the most challenging setting for the gradient norm approximation in classification.

\clearpage

\section{Regression Uncertainty}
\label{app:regression-figure}

\begin{figure}[h!]
    \centering
    \begin{subfigure}[t]{0.235\textwidth}
        \centering
        \includegraphics[width=\textwidth]{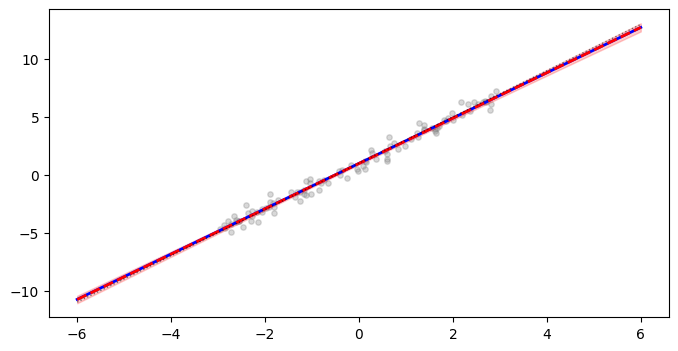}
        \caption{Linear, data}
    \end{subfigure}\hfill
    \begin{subfigure}[t]{0.235\textwidth}
        \centering
        \includegraphics[width=\textwidth]{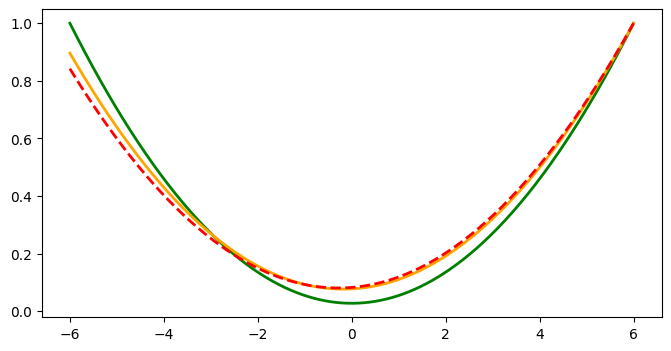}
        \caption{Linear, uncertainty}
    \end{subfigure}\hfill
    \begin{subfigure}[t]{0.235\textwidth}
        \centering
        \includegraphics[width=\textwidth]{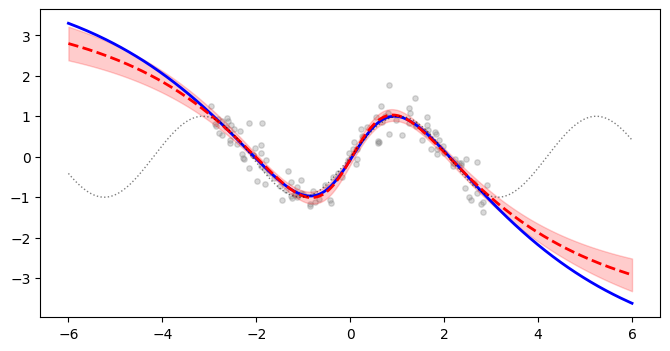}
        \caption{Nonlinear, data}
    \end{subfigure}\hfill
    \begin{subfigure}[t]{0.235\textwidth}
        \centering
        \includegraphics[width=\textwidth]{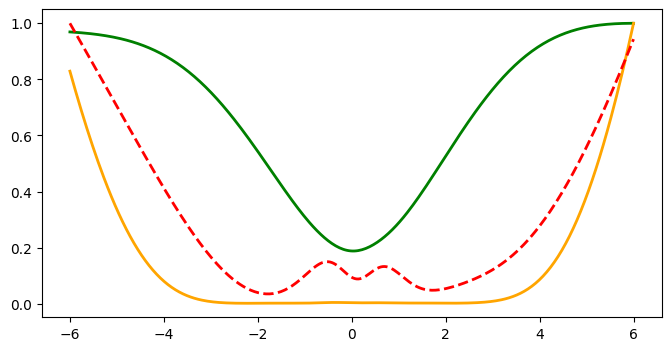}
        \caption{Nonlinear, uncertainty}
    \end{subfigure}
    \caption{Regression problems (columns 1 and 3) and normalized epistemic uncertainty (columns 2 and 4). In the data plots: gray dots are training samples, the dotted curve is the true data-generating function, the red dashed line is the MAP prediction, and the shaded band is the MCMC posterior predictive interval. In the uncertainty plots: the green solid line is $\|g\|^2$, the orange solid line is the Laplace approximation, and the red dashed line is the MCMC, all normalized to $[0, 1]$.}
    \label{fig:regression-overlay}
\end{figure}

\section{Validation Experiment Details}
\label{app:validation-details}

This appendix provides additional experimental details and per-problem results for the synthetic validation experiments in \cref{sec:experiments:validation}.

\subsection{Setup}

\paragraph{Binary classification.}
We train logistic regression and two-hidden-layer MLPs (with $\tanh$ activations) on three 2D binary classification problems: a linearly separable boundary, an XOR pattern, and concentric rings.
These problems span a range of decision boundary complexity, from a setting where the linear model is correctly specified to ones requiring nonlinear capacity.

\paragraph{Multiclass classification.}
We use softmax regression and two-hidden-layer MLPs on three 2D datasets: well-separated Gaussian clusters (3 classes), interleaved spirals (3 classes), and concentric rings (3 classes).
For multiclass models, we evaluate per-class epistemic uncertainty $\operatorname{Var}_\theta[p(y_c \mid x, \theta)]$ for each class $c$ and report correlations aggregated across classes.

\paragraph{Regression.}
We use a single-hidden-layer MLP with $\tanh$ activations (97 parameters) on two 1D problems: a linear function and a nonlinear function.

\paragraph{Scaling.}
We train a sequence of models of increasing width on the binary concentric rings problem, ranging from logistic regression (12 parameters) to a two-hidden-layer MLP with $1{,}028$ units per layer (approximately $1.07 \times 10^6$ parameters).

\subsection{Additional Results}

\subsubsection{Binary Classification}

\cref{tab:binary-epistemic} reports the full epistemic correlation results for binary classification, including gradient norm (GN), Laplace approximation (LA), and the correlation between the two.
On all three problems the GN--LA correlation exceeds $0.97$, meaning the Hessian correction provides almost no additional information beyond what the gradient norm already captures.
This is consistent with the role of the sigmoid nonlinearity discussed in \cref{sec:experiments:validation}: the output compression attenuates the gradient components that would otherwise expose posterior anisotropy, so the identity and the inverse Hessian produce nearly identical uncertainty maps.

\begin{figure}[h!]
    \centering
    \small
    \newcommand{\rowlabel}[1]{\rotatebox{90}{\parbox{2.2cm}{\centering\small #1}}}
    \begin{tabular}{c@{\hskip 2pt}ccc}
        & \textbf{MCMC} & \textbf{Gradient norm} ($\|g\|^2$) & \textbf{Laplace} ($g^\top H^{-1} g$) \\[2pt]
        \rowlabel{Linear (LogReg)} &
        \begin{subfigure}[t]{0.29\textwidth}
            \includegraphics[width=\textwidth]{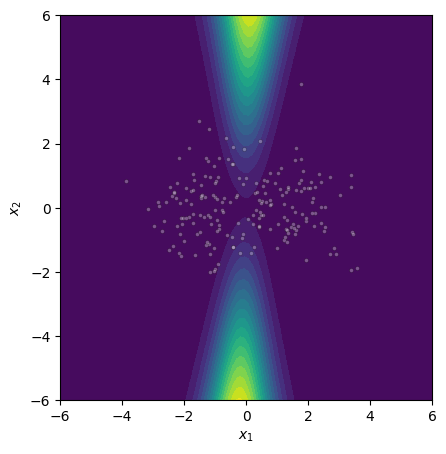}
        \end{subfigure} &
        \begin{subfigure}[t]{0.29\textwidth}
            \includegraphics[width=\textwidth]{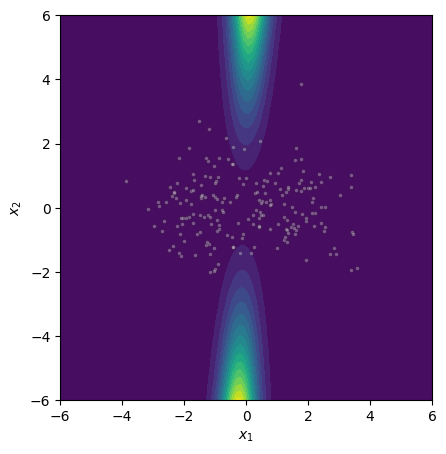}
        \end{subfigure} &
        \begin{subfigure}[t]{0.29\textwidth}
            \includegraphics[width=\textwidth]{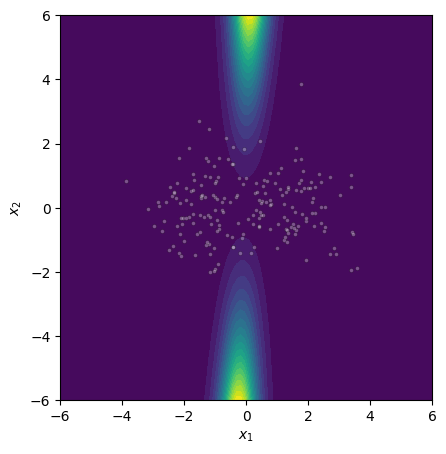}
            \phantomsubcaption\label{fig:binary-linear}
        \end{subfigure} \\[4pt]
        \rowlabel{XOR (MLP)} &
        \begin{subfigure}[t]{0.29\textwidth}
            \includegraphics[width=\textwidth]{figures/synthetic/binary/07_2d_xor_mc.png}
        \end{subfigure} &
        \begin{subfigure}[t]{0.29\textwidth}
            \includegraphics[width=\textwidth]{figures/synthetic/binary/07_2d_xor_gn.png}
        \end{subfigure} &
        \begin{subfigure}[t]{0.29\textwidth}
            \includegraphics[width=\textwidth]{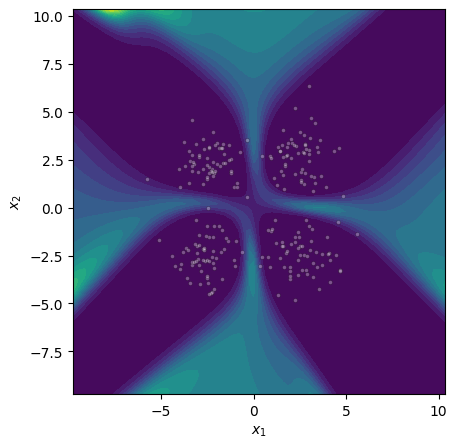}
            \phantomsubcaption\label{fig:binary-xor}
        \end{subfigure} \\[4pt]
        \rowlabel{Rings (MLP)} &
        \begin{subfigure}[t]{0.29\textwidth}
            \includegraphics[width=\textwidth]{figures/synthetic/binary/07_2d_mlp_mc.png}
        \end{subfigure} &
        \begin{subfigure}[t]{0.29\textwidth}
            \includegraphics[width=\textwidth]{figures/synthetic/binary/07_2d_mlp_gn.png}
        \end{subfigure} &
        \begin{subfigure}[t]{0.29\textwidth}
            \includegraphics[width=\textwidth]{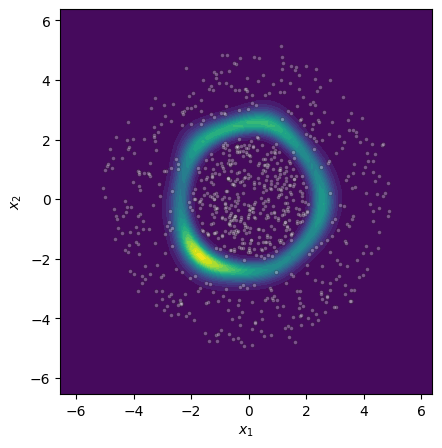}
            \phantomsubcaption\label{fig:binary-rings}
        \end{subfigure} \\
    \end{tabular}
    \caption{Epistemic uncertainty for all three binary classification problems. Each row shows a different problem; columns show MCMC, gradient norm, and Laplace approximation. On the linear problem all three methods are nearly identical; on XOR and rings the gradient norm and Laplace both recover the correct structure, while the Laplace approximation degrades on the nonlinear MLP problems.}
    \label{fig:binary-epistemic-all}
\end{figure}

\cref{fig:binary-xor} shows the epistemic uncertainty maps for the binary XOR problem, and \cref{fig:binary-linear} for the linear problem.

\begin{table}[h!]
    \centering
    \small
    \begin{tabular}{lcccccc}
        \toprule
        & \multicolumn{2}{c}{GN vs MCMC} & \multicolumn{2}{c}{LA vs MCMC} & \multicolumn{2}{c}{GN vs LA} \\
        \cmidrule(lr){2-3} \cmidrule(lr){4-5} \cmidrule(lr){6-7}
        Problem & $r$ & $\rho$ & $r$ & $\rho$ & $r$ & $\rho$ \\
        \midrule
        Linear (LogReg) & 0.95 & 0.99 & 0.95 & 0.99 & 1.00 & 1.00 \\
        XOR (MLP)       & 0.65 & 0.68 & 0.68 & 0.70 & 0.94 & 0.98 \\
        Rings (MLP)     & 0.86 & 0.44 & 0.86 & 0.46 & 0.97 & 1.00 \\
        \bottomrule
    \end{tabular}
    \caption{Binary classification: Pearson ($r$) and Spearman ($\rho$) correlation between epistemic uncertainty estimates and MCMC estimates. GN: gradient norm; LA: Laplace approximation; GN--LA: correlation between gradient norm and Laplace.}
    \label{tab:binary-epistemic}
\end{table}

\cref{tab:binary-aleatoric} reports aleatoric correlations.
The point estimate $p(y_c \mid x, \theta^*)(1 - p(y_c \mid x, \theta^*))$ tracks the MCMC estimates well on the linear problem ($r = 0.99$) and on the rings problem ($r = 0.95$), but less so on XOR ($r = 0.76$).
The Laplace-based aleatoric estimate performs poorly on the MLP problems, with negative correlations on XOR, suggesting that the Laplace posterior is a poor approximation to the true posterior in these settings.

\begin{table}[h!]
    \centering
    \small
    \begin{tabular}{lcccccc}
        \toprule
        & \multicolumn{2}{c}{PE vs MCMC} & \multicolumn{2}{c}{LA vs MCMC} & \multicolumn{2}{c}{PE vs LA} \\
        \cmidrule(lr){2-3} \cmidrule(lr){4-5} \cmidrule(lr){6-7}
        Problem & $r$ & $\rho$ & $r$ & $\rho$ & $r$ & $\rho$ \\
        \midrule
        Linear (LogReg) & 0.99 & 1.00 & 0.92 & 0.97 & 0.88 & 0.95 \\
        XOR (MLP)       & 0.76 & 0.74 & 0.08 & 0.12 & $-$0.10 & $-$0.11 \\
        Rings (MLP)     & 0.95 & 0.58 & 0.19 & 0.11 & 0.19 & 0.10 \\
        \bottomrule
    \end{tabular}
    \caption{Binary classification: aleatoric uncertainty correlations. PE: point estimate at MAP; LA: Laplace posterior mean. Both compared against MCMC posterior mean of $p(y_c \mid x, \theta)(1 - p(y_c \mid x, \theta))$.}
    \label{tab:binary-aleatoric}
\end{table}

\clearpage

\subsubsection{Multiclass Classification}

\cref{tab:multiclass-perclass} reports per-class epistemic correlations for the three multiclass problems; correlations are consistent across classes within each problem, confirming that the aggregate results are not driven by averaging over heterogeneous per-class performance.
\cref{tab:multiclass-aleatoric} reports per-class aleatoric correlations; the point estimate achieves consistently high correlations ($r \geq 0.95$, $\rho \geq 0.89$) across all problems and classes.

\begin{table}[h!]
    \centering
    \small
    \begin{minipage}[t]{0.48\textwidth}
        \centering
        \begin{tabular}{llcc}
            \toprule
            Problem & Class & $r$ & $\rho$ \\
            \midrule
            \multirow{5}{*}{Clusters (Softmax)}
            & 0       & 0.88 & 0.98 \\
            & 1       & 0.86 & 0.96 \\
            & 2       & 0.85 & 0.96 \\
            & 3       & 0.87 & 0.98 \\
            & Overall & 0.86 & 0.97 \\
            \midrule
            \multirow{5}{*}{Spirals (MLP)}
            & 0       & 0.82 & 0.92 \\
            & 1       & 0.79 & 0.91 \\
            & 2       & 0.72 & 0.88 \\
            & 3       & 0.72 & 0.93 \\
            & Overall & 0.76 & 0.91 \\
            \midrule
            \multirow{5}{*}{Rings (MLP)}
            & 0       & 0.92 & 0.88 \\
            & 1       & 0.90 & 0.96 \\
            & 2       & 0.88 & 0.95 \\
            & 3       & 0.89 & 0.95 \\
            & Overall & 0.88 & 0.97 \\
            \bottomrule
        \end{tabular}
        \caption{Multiclass: per-class Pearson ($r$) and Spearman ($\rho$) correlation between gradient norm and MCMC epistemic uncertainty.}
        \label{tab:multiclass-perclass}
    \end{minipage}
    \hfill
    \begin{minipage}[t]{0.48\textwidth}
        \centering
        \begin{tabular}{llcc}
            \toprule
            Problem & Class & $r$ & $\rho$ \\
            \midrule
            \multirow{5}{*}{Clusters (Softmax)}
            & 0       & 0.95 & 1.00 \\
            & 1       & 0.95 & 0.99 \\
            & 2       & 0.95 & 0.99 \\
            & 3       & 0.95 & 0.99 \\
            & Overall & 0.95 & 0.99 \\
            \midrule
            \multirow{5}{*}{Spirals (MLP)}
            & 0       & 0.96 & 0.98 \\
            & 1       & 0.95 & 0.98 \\
            & 2       & 0.97 & 0.96 \\
            & 3       & 0.96 & 0.97 \\
            & Overall & 0.96 & 0.97 \\
            \midrule
            \multirow{5}{*}{Rings (MLP)}
            & 0       & 0.99 & 0.89 \\
            & 1       & 0.98 & 0.97 \\
            & 2       & 0.95 & 0.97 \\
            & 3       & 0.95 & 0.97 \\
            & Overall & 0.96 & 0.98 \\
            \bottomrule
        \end{tabular}
        \caption{Multiclass: per-class Pearson ($r$) and Spearman ($\rho$) correlation between point-estimate and MCMC aleatoric uncertainty.}
        \label{tab:multiclass-aleatoric}
    \end{minipage}
\end{table}

\subsubsection{Regression}

\cref{tab:regression-details} reports the full regression results, including Hessian eigenvalue ranges that illustrate the degree of posterior anisotropy.
On the linear problem (2 parameters), the Hessian eigenvalues span a factor of $3.2\times$, and all three methods---gradient norm, Laplace, and MCMC---nearly coincide ($r \geq 0.98$).
On the nonlinear problem (97 parameters), the eigenvalue range spans a factor of $1.5 \times 10^4$, reflecting severe posterior anisotropy.
Here the Laplace approximation achieves $r = 0.93$ ($\rho = 0.97$) while the gradient norm drops to $r = 0.73$ ($\rho = 0.81$), confirming that the isotropy assumption is the primary source of error.

\begin{table}[h!]
    \centering
    \small
    \begin{tabular}{lcccccc}
        \toprule
        & \multicolumn{2}{c}{GN vs MCMC} & \multicolumn{2}{c}{LA vs MCMC} & \multicolumn{2}{c}{Hessian eigenvalues} \\
        \cmidrule(lr){2-3} \cmidrule(lr){4-5} \cmidrule(lr){6-7}
        Problem & $r$ & $\rho$ & $r$ & $\rho$ & Range & Ratio \\
        \midrule
        Linear (2 params)     & 0.98 & 0.99 & 1.00 & 1.00 & $[395, 1281]$      & $3.2\times$ \\
        Nonlinear (97 params) & 0.73 & 0.81 & 0.93 & 0.97 & $[1.0, 14648]$ & $1.5 \times 10^4$ \\
        \bottomrule
    \end{tabular}
    \caption{Regression: epistemic uncertainty correlations and Hessian eigenvalue ranges. The eigenvalue ratio $\lambda_{\max}/\lambda_{\min}$ quantifies posterior anisotropy.}
    \label{tab:regression-details}
\end{table}

\subsubsection{Scaling}

\cref{tab:scaling-full} reports the full scaling results for all nine model sizes.
The epistemic Spearman correlation follows the U-shaped trajectory described in \cref{sec:experiments:validation}, reaching a minimum at intermediate scales before recovering at the largest model sizes.

\begin{table}[h!]
    \centering
    \small
    \begin{tabular}{lrcccc}
        \toprule
        & & \multicolumn{2}{c}{Epistemic} & \multicolumn{2}{c}{Aleatoric} \\
        \cmidrule(lr){3-4} \cmidrule(lr){5-6}
        Model & $D$ & $r$ & $\rho$ & $r$ & $\rho$ \\
        \midrule
        LogReg             & 12         & 0.86 & 0.97 & 0.95 & 0.99 \\
        MLP(8,8)           & 132        & 0.51 & 0.84 & 0.83 & 0.87 \\
        MLP(16,16)         & 388        & 0.48 & 0.89 & 0.83 & 0.92 \\
        MLP(32,32)         & 1\,284     & 0.43 & 0.68 & 0.74 & 0.81 \\
        MLP(64,64)         & 4\,612     & 0.49 & 0.82 & 0.76 & 0.90 \\
        MLP(128,128)       & 17\,412    & 0.60 & 0.66 & 0.79 & 0.84 \\
        MLP(256,256)       & 67\,588    & 0.68 & 0.78 & 0.82 & 0.89 \\
        MLP(512,512)       & 266\,244   & 0.65 & 0.83 & 0.88 & 0.89 \\
        MLP(1028,1028)     & 1\,065\,012 & 0.74 & 0.87 & 0.89 & 0.93 \\
        \bottomrule
    \end{tabular}
    \caption{Scaling study: correlation between gradient norm estimates and MCMC estimates as a function of model size (number of parameters $D$) on the concentric rings problem.}
    \label{tab:scaling-full}
\end{table}

\subsubsection{Scaling with Mean-Field Variational Inference}
\label{app:scaling-vi}

To extend the scaling study beyond the regime where HMC is tractable, we repeat it using mean-field variational inference (VI) as the reference, which scales to substantially larger models.
We train MLPs with two hidden layers of equal width on a Gaussian-cluster classification problem ($N = 4{,}000$ training samples, four classes), sweeping width from 8 to 8\,192 (parameter count $D$ from $132$ to ${\sim}10^8$).
For each model we fit a diagonal Gaussian variational posterior via stochastic VI \citep{bingham-2019-pyrodeep}, and compute the gradient norm estimate at the variational mean.
\cref{tab:scaling-vi} reports Pearson and Spearman correlations between $\|g\|^2$ and the VI predictive variance, and between $p^*(1-p^*)$ and the VI aleatoric estimate.

\begin{table}[h!]
    \centering
    \small
    \begin{tabular}{lrcccc}
        \toprule
        & & \multicolumn{2}{c}{Epistemic} & \multicolumn{2}{c}{Aleatoric} \\
        \cmidrule(lr){3-4} \cmidrule(lr){5-6}
        Model & $D$ & $r$ & $\rho$ & $r$ & $\rho$ \\
        \midrule
        MLP(8,8)         &        132 & 0.79 & 0.96 & 0.94 & 0.99 \\
        MLP(16,16)       &        388 & 0.71 & 0.97 & 0.87 & 0.97 \\
        MLP(32,32)       &      1\,284 & 0.64 & 0.97 & 0.81 & 0.98 \\
        MLP(64,64)       &      4\,612 & 0.66 & 0.98 & 0.79 & 0.98 \\
        MLP(128,128)     &     17\,412 & 0.67 & 0.98 & 0.79 & 0.98 \\
        MLP(256,256)     &     67\,588 & 0.72 & 0.98 & 0.80 & 0.99 \\
        MLP(512,512)     &    266\,244 & 0.79 & 0.99 & 0.86 & 0.99 \\
        MLP(1028,1028)   &  1\,065\,012 & 0.76 & 0.98 & 0.83 & 0.98 \\
        MLP(2048,2048)   &  4\,210\,692 & 0.75 & 0.98 & 0.81 & 0.99 \\
        MLP(4096,4096)   & 16\,809\,988 & 0.76 & 0.98 & 0.82 & 0.99 \\
        MLP(8192,8192)   & 67\,174\,404 & 0.70 & 0.98 & 0.79 & 0.99 \\
        \bottomrule
    \end{tabular}
    \caption{Scaling study against a mean-field VI reference. With $N = 4{,}000$ training samples, the interpolation threshold falls between MLP(32,32) and MLP(64,64). The U-shape is visible in Pearson $r$, with the minimum at $D \approx N$, while Spearman $\rho$ stays above $0.96$ throughout.}
    \label{tab:scaling-vi}
\end{table}

The same U-shape appears in Pearson $r$, with the minimum at $D \approx N$ matching the MCMC study's interpolation threshold.
That the same pattern appears with two independent reference methods suggests it reflects a property of the approximation rather than an artifact of either reference.
Spearman $\rho$ stays above $0.96$ throughout, so the gradient norm preserves the uncertainty ranking even where the linear correlation dips.

\section{Question Answering Experiment Details}
\label{app:qa-details}

This appendix provides the full experimental details for the downstream question answering experiments in \cref{sec:experiments:downstream}.

\subsection{Models}

We evaluate four language models spanning a range of architectures and scales:
\begin{itemize}
    \item Llama 2 7B \citep{touvron-2023-llama2}, using a pre-quantized AWQ variant \citep{lin-2024-awqactivationaware, thebloke-2023-llama27bchatawq};
    \item Llama 3.2 3B \citep{grattafiori-2024-llama3};
    \item OLMo 1B \citep{groeneveld-2024-olmoaccelerating};
    \item Phi-4 \citep{abdin-2024-phi4technical}.
\end{itemize}
All models except Llama 2 are quantized to 4-bit precision using bitsandbytes \citep{dettmers-2023-qloraefficient} for computational efficiency.

\subsection{Baselines}

Following \citet{farquhar-2024-detectinghallucinations}, we compare against three baselines:
\begin{itemize}
    \item \textbf{Na\"ive entropy} \citep{farquhar-2024-detectinghallucinations}: multiple completions are sampled and entropy is computed over the length-normalised token-sequence log-probabilities, treating lexically distinct but semantically equivalent outputs as different.
    \item \textbf{Semantic entropy} \citep{kuhn-2022-semanticuncertainty}: multiple completions are sampled, clustered by semantic equivalence, and entropy is computed over the cluster probabilities.
    \item \textbf{P(True)} \citep{kadavath-2022-languagemodels}: the model is prompted to assess whether its own answer is correct, and the probability assigned to ``True'' is used as a confidence score.
\end{itemize}

\subsection{Evaluation}

Correctness is determined using the semantic equivalence criterion of \citet{farquhar-2024-detectinghallucinations}: an LLM judge checks whether the generated answer means the same as one of the reference answers in the context of the question.
For each uncertainty method, we train a logistic regression classifier to predict correctness from the uncertainty score and report the AUROC over 300 bootstrap runs per model--dataset configuration.
For the combined estimate (Epi.\ \& Alea.), the logistic regression uses both scores as features, making it a two-feature model versus one feature for all other methods.

\subsection{Datasets}

\paragraph{TriviaQA} \citep{joshi-2017-triviaqalarge} tests factual recall with unambiguous answers, e.g.\ ``What was the name of Michael Jackson's autobiography written in 1988?''
A model may be highly confident in a correct answer (it has memorized the fact) or highly confident in a wrong one (it has memorized a distortion), so uncertainty and correctness are largely independent.

\paragraph{TruthfulQA} \citep{lin-2022-truthfulqameasuring} targets common misconceptions and genuinely open questions, e.g.\ ``What happens to you if you eat watermelon seeds?''
The popular answer (they grow in your stomach) is false, while the set of accepted truthful answers is broad, ranging from ``nothing happens'' to ``they pass through your digestive system'', reflecting genuine variability in how the question can be correctly addressed.
The model therefore faces both epistemic conflict between what is commonly said and what is factually correct, and inherent ambiguity in what constitutes a complete answer.

\subsection{Per-Model Results}
\label{app:qa-per-model}

\begin{table}[h]
    \centering
    \small
    \begin{tabular}{lcccccc}
        \toprule
        & \multicolumn{6}{c}{TruthfulQA} \\
        \cmidrule(lr){2-7}
        Model & Na\"ive Ent. & Sem.\ Ent. & P(True) & Alea. & Epi. & Epi. \& Alea. \\
        \midrule
        Llama 2 (AWQ) & $0.57$ & $0.52$ & $0.57$ & $\mathbf{0.61}$ & $0.51$ & $0.58$ \\
        Llama 3.2 3B  & $0.53$ & $0.58$ & $0.50$ & $\mathbf{0.69}$ & $0.53$ & $0.68$ \\
        OLMo 1B       & $0.47$ & $0.47$ & $0.54$ & $0.51$ & $0.53$ & $\mathbf{0.57}$ \\
        Phi-4 (4-bit) & $0.47$ & $0.58$ & $0.58$ & $0.59$ & $0.63$ & $\mathbf{0.69}$ \\
        \midrule
        & \multicolumn{6}{c}{TriviaQA} \\
        \cmidrule(lr){2-7}
        Model & Na\"ive Ent. & Sem.\ Ent. & P(True) & Alea. & Epi. & Epi. \& Alea. \\
        \midrule
        Llama 2 (AWQ) & $0.58$ & $0.56$ & $\mathbf{0.75}$ & $0.61$ & $0.60$ & $0.61$ \\
        Llama 3.2 3B  & $0.53$ & $0.51$ & $0.59$ & $0.64$ & $0.51$ & $\mathbf{0.67}$ \\
        OLMo 1B       & $0.48$ & $0.58$ & $\mathbf{0.72}$ & $0.55$ & $0.55$ & $0.55$ \\
        Phi-4 (4-bit) & $0.50$ & $0.55$ & $\mathbf{0.69}$ & $0.61$ & $0.42$ & $0.60$ \\
        \bottomrule
    \end{tabular}
    \caption{Per-model AUROC for all methods (mean over 300 bootstrap runs). Best per row in bold. On TruthfulQA, at least one uncertainty estimate matches or exceeds all baselines for every model; on TriviaQA, baselines dominate for three of four models.}
    \label{tab:qa-per-model}
\end{table}

\subsection{Computational Complexity}
\label{app:complexity}

\cref{tab:timing-full} reports the per-model wall-clock timings summarized in \cref{sec:experiments:downstream}.
The corresponding per-sample computational cost beyond the shared answer-generation step is as follows: the entropy methods sample $K$ alternative completions to estimate predictive entropy; P(True) samples $K$ alternative completions and additionally runs one forward pass on a meta-prompt that includes the question, the original answer, and the alternatives; our method runs a single backward pass on the (already-generated) sequence to compute the gradient.

\begin{table}[h!]
    \centering
    \small
    \begin{tabular}{lccc}
        \toprule
        Method & Generations & Forward & Backward \\
        \midrule
        Gradient norm (ours) & 0 & 0 & 1 \\
        Na\"ive Entropy      & $K$ & 0 & 0 \\
        Semantic Entropy     & $K$ & 0 & 0 \\
        P(True)              & $K$ & 1 & 0 \\
        \bottomrule
    \end{tabular}
    \caption{Per-sample passes after the shared answer-generation step. ``Generations'' refers to additional autoregressive sampling runs (each producing multiple tokens). $K$ is the number of sampled alternative completions ($K = 10$ for the entropy methods; $K = 5$ for P(True)).}
    \label{tab:complexity}
\end{table}

\begin{table}[h!]
    \centering
    \small
    \begin{tabular}{lcccc}
        \toprule
        Model & Gradient norm (ours) & P(True) & Na\"ive Ent. & Sem.\ Ent. \\
        \midrule
        Llama 2 (AWQ) & $0.12 \pm 0.01$ & $7.14 \pm 0.96$ & $10.67 \pm 1.87$ & $11.29 \pm 2.03$ \\
        Llama 3.2 3B  & $0.08 \pm 0.00$ & $5.37 \pm 0.39$ & $8.18 \pm 0.77$  & $8.94 \pm 0.76$ \\
        OLMo 1B       & $0.06 \pm 0.00$ & $3.06 \pm 0.64$ & $4.52 \pm 1.08$  & $5.24 \pm 1.18$ \\
        Phi-4 (4-bit) & $0.15 \pm 0.00$ & $6.73 \pm 0.98$ & $10.26 \pm 1.50$ & $10.93 \pm 1.67$ \\
        \bottomrule
    \end{tabular}
    \caption{Wall-clock time per sample (mean $\pm$ std in seconds) on a single NVIDIA H100 GPU, excluding shared generation cost.}
    \label{tab:timing-full}
\end{table}

\subsection{Statistical Significance}
\label{app:qa-significance}

To assess whether the AUROC differences in \cref{tab:qa-results} are statistically significant, we perform paired $t$-tests across 10 random train/test splits (80/20), with AUROC values averaged over models within each split so that the 10 splits provide paired observations.
With 18 simultaneous tests (3 gradient methods $\times$ 3 baselines $\times$ 2 datasets), running each at $\alpha = 0.05$ would be expected to produce roughly one false positive by chance; we therefore apply the Benjamini--Hochberg (BH) procedure \citep{benjamini-1995-controllingfalse}, which adjusts each threshold to limit the expected \emph{fraction} of false discoveries among those declared significant, rather than requiring that every single test be free of error.
\cref{tab:qa-significance} reports BH-corrected $p$-values; bold entries are significant, and arrows indicate whether the gradient-based method is better ($\uparrow$) or worse ($\downarrow$) than the baseline.

\begin{table}[h]
    \centering
    \small
    \begin{tabular}{l ccc}
        \toprule
        & vs Na\"ive Ent. & vs Sem.\ Ent. & vs P(True) \\
        \midrule
        \multicolumn{4}{l}{\textit{TruthfulQA}} \\
        Aleatoric       & $\mathbf{<.001}\,\uparrow$ & $\mathbf{.005}\,\uparrow$  & $\mathbf{.005}\,\uparrow$ \\
        Epistemic       & $\mathbf{.005}\,\uparrow$  & $.575$                      & $.985$ \\
        Epi.\ \& Alea.\  & $\mathbf{<.001}\,\uparrow$ & $\mathbf{<.001}\,\uparrow$ & $\mathbf{<.001}\,\uparrow$ \\
        \midrule
        \multicolumn{4}{l}{\textit{TriviaQA}} \\
        Aleatoric       & $\mathbf{<.001}\,\uparrow$   & $\mathbf{<.001}\,\uparrow$   & $\mathbf{<.001}\,\downarrow$ \\
        Epistemic       & $.712$                         & $\mathbf{<.001}\,\downarrow$ & $\mathbf{<.001}\,\downarrow$ \\
        Epi.\ \& Alea.\  & $\mathbf{<.001}\,\uparrow$   & $\mathbf{<.001}\,\uparrow$   & $\mathbf{<.001}\,\downarrow$ \\
        \bottomrule
    \end{tabular}
    \caption{BH-corrected $p$-values from paired $t$-tests comparing each gradient-based method against each baseline, per benchmark. Bold = significant at $\alpha = 0.05$; $\uparrow$ = gradient method better, $\downarrow$ = baseline better. Tests use the Farquhar correctness criterion, 10 random splits, AUROC averaged over 4 LLMs per split.}
    \label{tab:qa-significance}
\end{table}

On TruthfulQA, the combined estimate significantly outperforms all three baselines ($p < 0.01$), as does the aleatoric estimate alone.
The epistemic estimate significantly exceeds na\"ive entropy ($p = .005$) but is statistically indistinguishable from semantic entropy and P(True).
On TriviaQA, P(True) significantly outperforms all gradient-based methods ($p < 0.001$); the epistemic estimate is significantly \emph{worse} than semantic entropy ($p < 0.001$) and indistinguishable from na\"ive entropy.

A paired bootstrap test (10\,000 resamples, paired by model and split) on the key comparison---Epi. \& Alea.\ vs.\ Aleatoric---yields $\Delta\text{AUROC} = {+}0.027$, 95\% CI $[0.002, 0.054]$, $p = 0.018$ on TruthfulQA, confirming that the epistemic term provides a statistically significant lift on this benchmark.
On TriviaQA, the lift is not significant ($\Delta\text{AUROC} = {+}0.006$, 95\% CI $[-0.003, 0.016]$, $p = 0.106$).

\subsection{Correlation Between Epistemic Uncertainty and P(True)}
\label{app:qa-correlation}

To assess whether the gradient-based epistemic estimate and P(True) capture redundant or complementary signal, we compute Spearman rank correlations on the raw per-sample values (\cref{tab:qa-correlation}).

\begin{table}[h]
    \centering
    \small
    \begin{tabular}{llrcl}
        \toprule
        Model & Dataset & $n$ & $\rho$ & $p$ \\
        \midrule
        Llama 2 (AWQ) & TruthfulQA & 811      & $-0.17$ & ${<}\,.001$ \\
        Llama 3.2 3B  & TruthfulQA & 794      & $-0.11$ & $.002$ \\
        OLMo 1B       & TruthfulQA & 501      & $-0.12$ & $.007$ \\
        Phi-4 (4-bit) & TruthfulQA & 455      & $-0.20$ & ${<}\,.001$ \\
        Llama 2 (AWQ) & TriviaQA   & 7\,885   & $-0.22$ & ${<}\,.001$ \\
        Llama 3.2 3B  & TriviaQA   & 7\,408   & $-0.27$ & ${<}\,.001$ \\
        OLMo 1B       & TriviaQA   & 4\,670   & $-0.19$ & ${<}\,.001$ \\
        Phi-4 (4-bit) & TriviaQA   & 3\,270   & $-0.10$ & ${<}\,.001$ \\
        \midrule
        \textit{Pooled} & TruthfulQA & 2\,561  & $-0.23$ & ${<}\,.001$ \\
        \textit{Pooled} & TriviaQA   & 23\,233 & $-0.21$ & ${<}\,.001$ \\
        \bottomrule
    \end{tabular}
    \caption{Spearman rank correlation ($\rho$) between the epistemic uncertainty estimate ($\|g\|^2$) and P(True) on raw per-sample values.}
    \label{tab:qa-correlation}
\end{table}

All correlations are negative and significant: higher epistemic uncertainty (larger gradient norm) corresponds to lower self-assessed confidence, as expected.
However, the magnitudes are weak ($|\rho| \approx 0.10$--$0.27$), indicating approximately 4\% shared variance at the pooled level.
This confirms that the gradient-based estimate captures information largely complementary to the model's self-assessed confidence, consistent with the observation that the two measures are most useful on different benchmarks (\cref{sec:experiments:downstream}).

\subsection{Cross-Model Transfer}
\label{app:qa-lomo}

To test whether the gradient-based epistemic estimate generalizes across architectures, we run a leave-one-model-out (LOMO) experiment: for each of the four LLMs, we train a logistic regression classifier on the remaining three and evaluate on the held-out model.
The raw $\|g\|^2$ depends on the absolute scale of the parameters, which varies across model families and quantization schemes; dividing by $\|\theta^*\|^2$ removes this dependence and should in principle yield a relative measure that is comparable across architectures.
We compare the raw $\|g\|^2$ against this parameter-norm-normalized variant ($\|g\|^2 / \|\theta^*\|^2$), both alone and combined with the aleatoric estimate.

\begin{table}[h]
    \centering
    \small
    \begin{tabular}{ll cccc}
        \toprule
        & Held-out & Raw & Normalized & Raw \& Alea.\ & Norm \& Alea.\ \\
        \midrule
        \multirow{5}{*}{TruthfulQA}
        & Llama 3.2 3B  & $0.53$ & $0.47$ & $0.33$ & $0.33$ \\
        & Llama 2 (AWQ) & $0.55$ & $0.55$ & $0.51$ & $0.53$ \\
        & OLMo 1B       & $0.47$ & $0.47$ & $0.49$ & $0.49$ \\
        & Phi-4 (4-bit) & $0.37$ & $0.37$ & $0.61$ & $0.61$ \\
        \cmidrule(lr){2-6}
        & \textit{Mean (Std)} & $0.48\,(0.08)$ & $0.46\,(0.08)$ & $0.48\,(0.12)$ & $0.49\,(0.12)$ \\
        \midrule
        \multirow{5}{*}{TriviaQA}
        & Llama 3.2 3B  & $0.49$ & $0.49$ & $0.35$ & $0.35$ \\
        & Llama 2 (AWQ) & $0.40$ & $0.40$ & $0.38$ & $0.38$ \\
        & OLMo 1B       & $0.45$ & $0.55$ & $0.44$ & $0.44$ \\
        & Phi-4 (4-bit) & $0.40$ & $0.40$ & $0.61$ & $0.61$ \\
        \cmidrule(lr){2-6}
        & \textit{Mean (Std)} & $0.44\,(0.05)$ & $0.46\,(0.07)$ & $0.45\,(0.11)$ & $0.45\,(0.11)$ \\
        \bottomrule
    \end{tabular}
    \caption{Leave-one-model-out AUROC: train on 3 models, evaluate on held-out 4th. Values are at or below chance ($0.50$), and the relationship occasionally inverts across architectures. Normalization by the parameter norm does not improve transfer.}
    \label{tab:qa-lomo}
\end{table}

All configurations produce chance-or-below AUROC on held-out models (\cref{tab:qa-lomo}), and the relationship occasionally inverts---Phi-4's raw epistemic score drops to $0.37$--$0.40$, meaning higher gradient norm predicts \emph{correct} answers when trained on other models---confirming that the mapping between gradient magnitude and correctness is architecture-specific with no consistent direction across models.
Parameter-norm normalization does not improve transfer: the mean AUROC and cross-model variance are essentially unchanged.
The one apparent exception is Phi-4 under the combined feature set ($0.61$ on both benchmarks), but inspection shows this lift comes entirely from the aleatoric term: the raw epistemic score alone is $0.37$--$0.40$ for Phi-4, below chance, while the aleatoric score transfers because it is architecture-agnostic (depending only on output probabilities, not gradient magnitudes).
The high cross-model standard deviation ($0.11$--$0.12$) for the combined features is driven by this single model; without Phi-4, all means drop further below chance.
This is consistent with the high per-model variance observed in \cref{tab:qa-per-model} and indicates that the epistemic estimate should be calibrated per model rather than applied with a universal threshold.

\end{document}